\documentclass{article}

\usepackage[preprint]{neurips_2024}

\usepackage[utf8]{inputenc} 
\usepackage[T1]{fontenc}    
\usepackage{hyperref}       
\usepackage{url}            
\usepackage{booktabs}       
\usepackage{amsfonts}       
\usepackage{nicefrac}       
\usepackage{microtype}      
\usepackage{xcolor}         
\usepackage{amsmath}
\usepackage{enumitem}
\usepackage{multirow}
\usepackage{array} 
\usepackage{ragged2e}
\usepackage{vcell}
\usepackage{float} 
\usepackage{graphicx}
\usepackage{subfigure}
\usepackage{graphicx}
\usepackage{url}

\title{Chain-of-Planned-Behaviour Workflow Elicits Few-Shot Mobility Generation in LLMs}

\author{%
  Chenyang~Shao  \\
  Tsinghua University\\
  Beijing, China \\
  \texttt{shaocy20@mails.tsinghua.edu.cn} \\
  \And
  Fengli~Xu
  \thanks{Corresponding authors.}\\
  Tsinghua University \\
  Beijing, China \\
  \texttt{fenglixu@tsinghua.edu.cn} \\
  \And
  Bingbing~Fan \\
  Tsinghua University \\
  Beijing, China \\
  \texttt{1441257405@qq.com} \\
  \And
  Jingtao~Ding \\
  Tsinghua University \\
  Beijing, China \\
  \texttt{dingjingtao@tsinghua.edu.cn} \\
  \And
  Yuan~Yuan \\
  Tsinghua University \\
  Beijing, China \\
  \texttt{y-yuan20@mails.tsinghua.edu.cn} \\
  \And
  Meng~Wang \\
  Hefei University of Technology \\
  Hefei, China \\
  \texttt{wangmeng@hfut.edu.cn} \\
  \And
  Yong~Li$^{*}$ \\
  Tsinghua University \\
  Beijing, China \\
  \texttt{liyong07@tsinghua.edu.cn} \\
}

\begin{document}

\maketitle

\begin{abstract} 
The powerful reasoning capabilities of large language models (LLMs) have brought revolutionary changes to many fields, but their performance in human behaviour generation has not yet been extensively explored.
This gap likely emerges because the internal processes governing behavioral intentions cannot be solely explained by abstract reasoning. Instead, they are also influenced by a multitude of factors, including social norms and personal preference.
Inspired by the \textit{Theory of Planned Behaviour} (TPB), we develop a LLM workflow named \underline{\textbf{C}}hain-\underline{\textbf{o}}f-\underline{\textbf{P}}lanned-\underline{\textbf{B}}ehaviour (CoPB) for mobility behaviour generation, which reflects the important spatio-temporal dynamics of human activities. Through exploiting the cognitive structures of \textit{attitude}, \textit{subjective norms}, and \textit{perceived behaviour control} in TPB, CoPB significantly enhance the ability of LLMs to reason the intention of next movement.
Specifically, CoPB substantially reduces the error rate of mobility intention generation from 57.8\% to 19.4\%. 
To improve the scalability of the proposed CoPB workflow, we further explore the synergy between LLMs and mechanistic models. We find mechanistic mobility models, such as gravity model, can effectively map mobility intentions to physical mobility behaviours. The strategy of integrating CoPB with gravity model can reduce the token cost by 97.7\% and achieve better performance simultaneously. Besides, the proposed CoPB workflow can facilitate GPT-4-turbo to automatically generate high quality labels for mobility behavior reasoning. We show such labels can be leveraged to fine-tune the smaller-scale, open source LLaMA 3-8B, which significantly reduces usage costs without sacrificing the quality of the generated behaviours. Our code and datasets are available: 
\href{https://anonymous.4open.science/r/CoPB}{https://anonymous.4open.science/r/CoPB}
  
\end{abstract}


\section{Introduction}

Human mobility behaviour is an important aspect of human activities, often revealing the nuanced details of human's spatio-temporal interaction with the physical environment~\cite{gonzalez2008understanding}. Acquisition of mobility behaviors plays a crucial role in numerous important applications, such as urban planning~\cite{xu2021emergence}, epidemic control~\cite{chen2022strategic} and business site selection~\cite{liu2017point}. Traditionally, large-scale human mobility behaviour is collected through household travel survey~\cite{stopher2007household} or ubiquitous personal sensors like smartphone~\cite{zheng2010geolife}, which can be hugely expensive~\cite{jiang2016timegeo} and are associated with serious privacy risks~\cite{de2015unique,xu2017trajectory}. Therefore, it has been a long-standing research quest to design low-cost yet high-quality generative models for human mobility data~\cite{song2010modelling,feng2020learning}.

Mechanistic models are an classical solution to mobility behaviour generation~\cite{song2010modelling, brockmann2006scaling}, which employ stochastic processes to describe the movement process and use few parameters to model mobility strategies.
With the advancement of deep generative networks, GANs~\cite{yuan2023learning,yuan2022activity,wang2023pategail}, VAEs~\cite{long2023VAE}, and Diffusion methods~\cite{zhu2023difftraj} have gradually emerged in this field.~\cite{luca2021survey} By learning from large datasets through gradient descent, deep neural networks can accurately capture distribution features, thus ensuring high fidelity in sampled data. 
Both mechanistic models and deep generative models can be summarized as paradigms where a data distribution is first fitted and then results are sampled from it. The scarcity of semantics and the demand for large amounts of data are common problems for both. 
Therefore, it remains a challenge to design accurate, sample-efficient and semantic-aware generative models for human mobility.

The rapid progress of large language models(LLMs) in the past two years has heralded fresh prospects for the domain of mobility generation.
The scaled-up versions of LLMs such as GPT~\cite{floridi2020gpt} and PaLM~\cite{anil2023palm} have shown emergent abilities for general purpose reasoning~\cite{wei2022emergent}. For example, research has discovered that explicitly guiding the thought process through prompts design can greatly unleash the inferential potential of LLMs.~\cite{wei2022chain, yao2023tree}. Besides, LLMs also exhibit impressive role play ability that allows them to simulate individuals with specific profiles~\cite{park2023generative}. These advancements and explorations seem to foreshadow the immense potential of deploying LLMs in mobility behaviour generation, yet they are juxtaposed with challenges. As the intrinsic factors controlling mobility behaviours cannot be simply explained through abstract reasoning. They are also influenced by personal preferences and the pressure exerted by subjective norms. At the same time, current objective conditions and potential difficulties encountered during implementation will affect people's perceived judgment of the feasibility of behaviour, thereby influencing the final intention decision.


To tackle these challenges, we draw insights from the \textit{Theory of Planned Behaviour}\cite{ajzen1991theory}, and design a novel behavioral intention reasoning workflow \textbf{Chain-of-Planned-Behaviours (CoPB)} from the intertwining of \textit{attitudes}, \textit{subjective norms}, and \textit{perceived behavioural control}. As this reasoning workflow is best suited for step-by-step intention reasoning, we decompose the complete mobility behaviour generation into a coupled decision chain that recursively generates the next behavioural intention.
Since \textit{attitudes} and \textit{subjective norms} are relatively static concepts, they only need to be queried once before the overall generation process. \textit{Perceived behavioral control}, on the other hand, will be dynamically updated based on the preceding intention sequence, influencing the likelihood assessment of the next intention. Therefore, at each generation step, the current perceived behavioral control is inferred first, followed by the inference of the next intention.
Additionally, a few manually annotated reasoning trajectories are included in the prompts to unlock the few-shot in-context learning capability of the LLM.
Evaluation on two real-world mobility datasets indicates that our workflow significantly enhances the quality of intention generation, notably reducing the error rate of mobility intention generation from 0.578 and 0.642 to 0.194 and 0.247, respectively.

To improve the scalability of the proposed CoPB workflow, we further explore the synergistic effects between LLM and mechanistic models. Utilizing a gravity model, we map the intentions generated by LLMs to physical space and generate reasonable choices based on geographical distance and density of Points of Interest (POIs), thus generating complete mobility behaviours. This strategy not only drastically reduces the token consumption but also achieves better generation quality in both behaviour intentions and physical trajectories. Specially, our workflow incurs only 1/40 of the token consumption per generated trajectory compared to pure LLM methods where geospatial information is all included in the prompts, and outperforms all deep generative models on all evaluation metrics.
Furthermore, building upon our proposed workflow, we constructed a Q\&A dataset for intention generation using GPT-4-turbo, and use it to fine-tune the LLaMA3-8B model\cite{LLaMA3}, resulting in LLaMA3-8B's mobility generation capability closely rivaling that of GPT-4-turbo. This significantly reduces the cost of generation even further, both in terms of time and token consumption. To summary, the contributions of our work can be summarized as follows:

\begin{itemize}
    \item We are the first to introduce the \textit{Theory of Planned Behaviour} into mobility behaviour generation with LLMs. Based on this theory, we have constructed an effective behavioral intention reasoning workflow.
    
    \item Through the synergistic effect between LLMs and mechanistic models, we reduce token cost by an order of magnitude while ensuring the high quality in generated trajectories.    
    
    \item Utilizing GPT-4-turbo, we automate the construction of mobility behaviour dataset and use it to fine-tune LLaMA 3-8B, achieving performance comparable to GPT-4-turbo. This significantly mitigates the cost of generation.
    
\end{itemize}

\section{Related Work}

\textbf{Deep Generative Models for Mobility Data.}
Deep generative models have been widely applied to human trajectory generation problems. Feng et al.~\cite{feng2020learning} incorporate an attention-based region network into the generator of an adversarial network to introduce prior information about urban structures. Long et al.~\cite{long2023VAE} design two VAEs to model the locations of users' homes and workplaces as well as their temporal patterns. Zhu et al.~\cite{zhu2023difftraj} embedding conditional information in the reverse denoising process in the diffusion model to improve the fidelity of trajectories. Fundamentally, deep generative models are aimed at capturing and replicating the complex distributions of real-world data. GANs imitate data through a powerful generator under the supervision of a discriminator. VAEs map input data to a latent space through an encoder, and then map latent variables back to the original data space through a decoder, maximizing the likelihood to ensure that the data distribution generated by the decoder is consistent with the real one. Meanwhile, diffusion models restore the original data step by step through a reverse denoising process from noise. The general process of these mothods can be summarized as follows: accurately model a distribution based on extensive data and subsequently generate samples from it, which significantly diverges from our paradigm.

\textbf{LLM for Behaviour Simulation.}
LLMs have increasingly shown their potential in the domain of behavioural generation, marking an innovative leap in how behavioural data can be acquired. Recent works ~\cite{qian2023communicative, hamalainen2023evaluating, park2023generative, argyle2023out, shanahan2023role, aher2023using} demonstrate the ability of LLMs to generate complex, nuanced survey questions and scenarios, potentially replacing traditional methods of survey creation and administration. At the heart of this capability is the LLMs' advanced role-play capability, which enables them to assume the perspectives of various respondents, thereby generating more dynamic and engaging survey content. 
This function not only allows for the creation of diverse and tailored survey materials but also opens the door to understanding intricate behavioural patterns through simulated interactions. However, despite these promising advancements, there is a growing concern regarding the inherent biases that may be perpetuated or even exacerbated by LLMs. \cite{hagendorff2023human, acerbi2023large, santurkar2023whose} As these models learn from vast datasets, they are susceptible to mirroring the biases present in their training data. This could lead to skewed question generation, potentially influencing the responses in ways that reinforce existing stereotypes or biases, thereby affecting the validity of the outcomes. To the best of our knowledge, this work is a pioneering attempt that introduces LLMs into mobility behaviour generation.

\section{Preliminaries}


\subsection{Problem Definition}

\textbf{\textit{Mobility behaviour.}} A piece of mobility behaviour data consists of a sequence of points coherent in time, donated by $x = \{ s_1, s_2,...,s_n \}$, where each point $s_i$ can be expressed as ($t_i$, $l_i$, $e_i$). $l_i$ denotes the spatial position, either in the form of region ID or coordinate of latitude and longitude. $t_i$ represents the start time and end time of the visit to $l_i$. $e_i$ represents the type of visit intention.

\textbf{\textit{Mobility behaviour generation task.}} Given a real-world mobility behaviour dataset $\mathcal{X} = \{x^1,x^2,...x^N \}$. The objective of the task is to generate new mobility behaviour data $\mathcal{Y} = \{y^1,y^2,...y^N \}$ such that the generated data exhibits similar characteristics and mobility patterns as the original real data.

\textbf{\textit{Deep generative models for mobility behaviour.}}
Deep generative models rely on deep neural networks to directly learn spatiotemporal patterns and fit data distributions from real data. Assume that the real data $x_0,x_1,...x_n$ satisfy an unknown data distribution $p_{data}(x)$ and the data generated by the deep generation model satisfy a distribution $p_{\theta}$. The optimization objective is to make $p_{\theta}$ and $p_{data}$ consistent. KL divergence is a commonly used method.
In practice, since $p_{data}(x)$ is often unknown, the objective can be approximated using methods such as \textit{Maximum Likelihood Estimation} or by minimizing alternative divergences or distances. Let's use $\widetilde{p}_{data}(x^{\prime} )$ to donate the generated data distribution after training, and then sampling process can be represented as $x^{\prime} \sim\widetilde{p}_{data}(x^{\prime})$

\textbf{\textit{Generation of large language models.}}
Let's first use $p_{\phi}$ to denote a pre-trained LLM with parameters $\phi$. Then use $u=(u[1],...,u[n])$ for input language sequence, and $v=(v[1],...,v[m])$ for output language sequence where each sequence is composed of numerous tokens. The generation process for $v$ can be represented as: $p_{\phi}(v)=\prod \limits_{i=1}^n p_\phi (v[i]|u,v[1..i-1])$. Next, let $u$ donate the input prompts, which can be global task information, few-shot examples, or notice for canonical outputs and let $v$ denotes the output answer. Then the Q\&A process can be simply represented as: $v\sim p_{\phi}(u)$. 
Unlike the fit-sample paradigm of deep generative models, using the reasoning capabilities of LLMs for generation represents a promising alternative, which can be applied to various generation tasks.

\section{Methodology}

\begin{figure*}[t]
    \centering
    \includegraphics[width=1\linewidth]{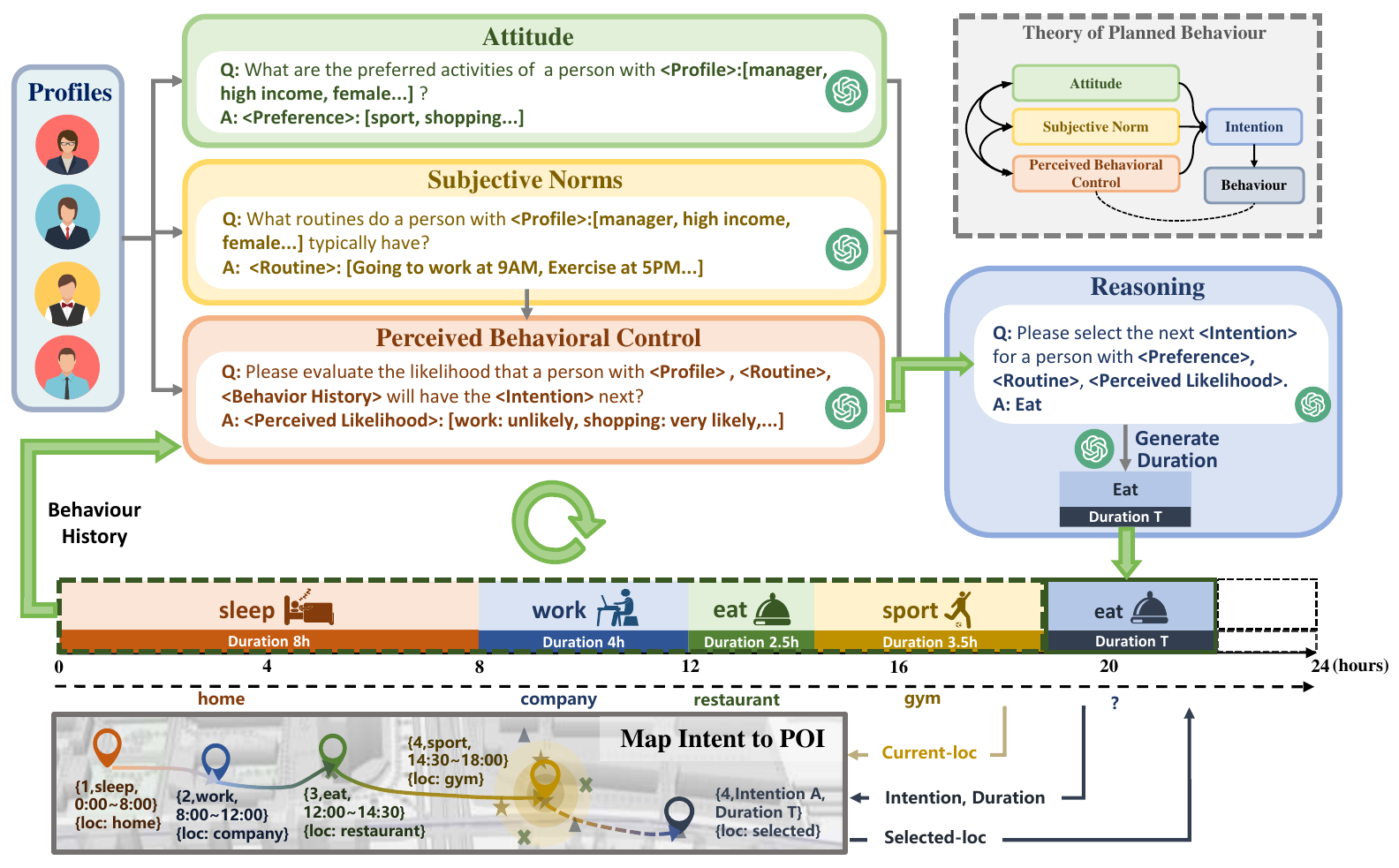}
    \vspace{-5mm}
    \caption{The overview of our proposed \textbf{CoPB} workflow. The bold green arrows indicate that CoPB will update perceived behavioral control and generate mobility behavior step-by-step.}
    \label{fig:system1}
\end{figure*}

As shown in Figure \ref{fig:system1}, our overall framework achieves the generation of abstract intentions and specific physical space movement behaviours. In the following subsections, we will first introduce the core part of reasoning—reshaping LLM reasoning using the \textit{Theory of Planned Behaviour} (\ref{method-TPB}). Next, we will present the overall workflow of intention generation (\ref{method-Workflow}). In subsection \ref{method-gravity}, we will explain how the mechanistic model collaborates with the LLM. Finally, we will describe the construction of the mobility dataset and how to fine-tune LLaMA3-8B in subsection \ref{method-llama3}.

\subsection{Theory of Planned Behaviour} \label{method-TPB}
The \textit{Theory of Planned Behaviour}\cite{ajzen1991theory} was developed by Icek Ajzen as a framework to comprehend and forecast behaviour across a wide range of different categories of behaviours in psychology and behavioural science, suggesting that actions are directly influenced by behavioural intentions and, in some cases, perceived behavioural control. These behavioural intentions are shaped by three elements: \textit{\textbf{attitudes toward behaviours, subjective norms, and perceived behavioural control}}, with each predictor weighted for its importance in relation to the behaviour and character profiles.

\begin{itemize}[leftmargin=*]
    \item  \textit{\textbf{Attitudes toward behaviours}} refer to an individual's positive or negative feelings toward intentions, stemming from the evaluation of both the intentions themselves and the results they may produce. This encompasses numerous subjective perspectives, intricately intertwined with the individual's cognitive depth, preferences, and life experiences, thus closely connected to the character's portrait attributes. In our implementation, we ask LLM agents about their preferences towards various intentions based on profiles, which can be donated as: $\alpha = \langle Preference \rangle: [preference_1, preference_2, ...]$
    
    \item \textit{\textbf{Subjective norms}} involve individuals' perceptions of whether they should engage in a particular behaviour, influenced by societal moral standards, current trends, public opinions, or the viewpoints of significant others such as family and friends. 
    Contemporarily, external influences from others or societal judgments often overshadow one's own stance, emerging as significant factors in shaping behaviour. 
    Simultaneously, they also compels individuals to adhere to expected norms or long-term planning criteria for their behaviours. In our implementation, we incorporate an additional query to let the LLM agent generate its own planning routines based on the profile, which can be represented as: $ \beta = \langle Routine \rangle: [routine_1, routine_2, ...]$
    
    \item \textit{\textbf{Perceived behavioural control}} refers to the extent to which an individual believes they can complete a specific behaviour successfully, influenced by the behaviour type, current conditions and obstacles encountered. This factor can be explained as a comprehensible intuition: \textit{Better safe than sorry}. When an individual feels highly confident about a particular behaviour, he is naturally more inclined to execute it. Based on this point, we explicitly instruct the LLM to evaluate the feasibility of the subsequent arrangements based on the preceding generation results, and provide a perceived likelihood assessment, which can be donated as: $\gamma = \langle Perceived \ likelihood\rangle: [intention_1:likelihood_1, intention_2:likelihood_2, ...]$

\end{itemize}





\subsection{Chain-of-Planned-Behaviour Workflow} \label{method-Workflow}

In this section, we will introduce the construction of the entire LLM workflow. The input to the LLM includes two parts: the specified user profile information and eight pieces of manually annotated in-context intention sequence examples where we have manually annotated the reasons why the individual make the next step intention decision and time arrangement. We can use $\mathbf{u}$ to represent these two kinds of information collectively. The expected output includes intention sequences containing intentions and detailed start and end times, donated as $\mathbf{v} = \{(t_0,e_0),...(t_n,e_n)\}$.  \emph{e.g.}, [["(00:00, 08:33)", "go to sleep"], ["(09:47, 17:49)", "go to work"], ["(18:45, 19:49)", "eat"], ["(20:01, 20:35)", "do shopping"], ["(21:40, 23:59)", "go home"]]


If we directly let the LLM generate $\mathbf{v}$ based on the input $\mathbf{u}$, the generated results are often too simple and stylized, far from the normal human life arrangements, indicating that this task is non-trivial. Therefore, we draw on the idea of COT, breaking down the complete $\mathbf{v}$ into multiple intention decision steps. Each time, through a query based on the input $\mathbf{u}$ and the preceding generation results $\mathbf{v_{i}} = \{ (t_0,e_0),...,(t_i,e_i)\}$, we obtain $e_{i+1}\sim p_{\phi}(\mathbf{v_i},\mathbf{u}) $, which is then incorporated into the inquiry to obtain $t_{i+1}\sim p_{\phi}(\mathbf{v_i},\mathbf{u}, e_{i+1}) $. Purpose of decomposition is to construct reachable intermediate steps between two distant input-output pairs $\mathbf{u}$ and $\mathbf{v}$. In this way, the process from $(t_i, e_i)$ to $(t_{i+1},e_{i+1})$ can become easier, more accurate, and more controllable. Therefore, LLM can confidently follow a promising behaviour path to reach reasonable generation results at the endpoint.

Furthermore, for each step $e_{i+1}\sim p_{\phi}(\mathbf{v_{i}},\mathbf{u}) $ of intention decision, we integrate the \textit{Theory of Planned Behaviour} into it. First, based on the profile($p$), we will ask the agent about its daily preferences ($\alpha$). Then, also based on the profile, we have the agent generate anchor points or routines in life ($\beta$). Finally, using profile, generated preferences and anchor points, combined with the preceding generation results, we ask the agent to give a dynamic perceived likelihood ($\gamma_i$) of the next intention. Finally, we combine $p$, $\alpha$, $\beta$, $\gamma_i$ as comprehensive inputs to generate $e_{i+1}$,  which can be donated as follows. The organization and design of prompts are indicated using $\mathtt{prompt}$.
\begin{equation} \label{QA12}
    \begin{split}
    \alpha \sim p_{\phi}(\mathtt{prompt_\alpha}(p))  \quad \beta \sim p_{\phi}(\mathtt{prompt_{\beta}}(p)) \quad
    \gamma_i\sim p_{\phi}(\mathtt{prompt_{\gamma}}(p, \mathbf{v_{i}},\alpha,\beta))  
    \end{split}
\end{equation}
\begin{equation} \label{QA3}
    \begin{aligned}
    e_{i+1}\sim p_{\phi}(\mathtt{prompt_{e}}(\mathbf{v_i}, \alpha, \beta, \gamma_{i}))  
    \end{aligned}
\end{equation}

The profile($\alpha$) and life anchor points($\beta$) serve as static contexts, influencing the overall decision process. While the perceived likelihood evaluation($\gamma$) for the next action as well as the temporal context reflected in the historical sequence is dynamic and constantly changing. This mechanism will guide the LLM to generate in a more realistic and reasonable direction.

Besides, we compile the distribution of the number of daily intentions from a small amount of real data (200 pieces of 7-day trajectories), from which we would sample as a limit on the total number of intentions to be generated before the generation process starts. The total of 10 intentions considered in our experiment are as follows: 1. go to work 2. go home 3. eat 4. do shopping 5. do sports 6. excursion 7. leisure or entertainment 8. sleep 9. medical treatment, 10. handle the trivialities of life. In fact, the granularity of intention type differentiation can be freely controlled according to semantic requirements.

\subsection{Mapping intentions to Behaviours with Mechanistic Model} 
\label{method-gravity}

In the previous part, we introduced how to construct the LLM workflow to generate intention sequences, while here we will introduce how to select suitable POI location for the intention, thereby mapping to actual mobility behaviour. The home places can be inferred from real trajectory data or sampled from demographic data, but how to correspond other abstract intentions to specific ones in millions of POI locations remains a challenge. 

Previous approaches either simplified location selection to choosing from regular geographic grids or generating the coordinates of latitude and longitude numerically. However, both approaches lacked realism and semantic richness, failing to match real POIs accurately. Here we decide to use the simple but efficient gravity model to handle this. 
The gravity model emphasizes the importance of distance in human migration. It considers that the magnitude of migrating traffic between two regions can be reduced to such a simple form: 
\begin{equation}
\begin{aligned}
    P_{i,j} \propto \frac {P_i P_j}{r_{ij}}
    \end{aligned}
\end{equation}
where $P_i$ and $P_j$ generally donate the population of region i and region j, respectively. And $r_{ij}$ donates the distance between region i and region j. Taking the nonlinear factors into account, it can be further expressed as: 
\begin{equation}
    \begin{aligned}
    P_{i,j} = K m_i m_j f(r_{ij})
    \end{aligned}
\end{equation}
where $m_i$ is the function of $P_i$ and is typically expressed in the form of $P_i^{\alpha}$, as is $m_j$. $f(r_{ij})$ represents a decreasing function of distance and $K$ is a normalization constant. All of these functions can be fitted. 
In our experiment, we replace the population num with the density of POIs. We believe that the higher the density of POIs, the more attractive they are to individuals. With the previous location as the center of the circle, the surrounding map is partitioned into concentric rings at 1km intervals. For all POIs in one ring, $m_j$ is set to the POI density in the ring (number/ring area), and $f(r_{ij})$ is set to $r_{ij}^{-2.5}$ which is fitted from 200 pieces of trajectories. We assume that the great majority of people travel an upper distance limit of 10km, so the POI search is limited to circles within a radius of 10km.


    
    

\subsection{Dataset Construction and LLaMA 3 Finetuning} \label{method-llama3}

Mobility generation model is often used for city-scale mobility generation tasks, which often require millions or even more trajectories to be generated. However, using GPT series models for mobility behaviour data generation will inevitably result in significant API consumption. 

Therefore, we decided to build a mobility behaviour dataset based on multi-turn Q\&A logs and finetune it on an open-source model to achieve the same performance in mobility behaviour generation. Q\&A logs need to include four most important inquiries in our workflow: (a) inquiring user preferences regarding attitude, (b) daily routine inquiries for subjective norm, (c) perceived likelihood of the next intention for perceived behaviour control, (d) asking the final intention decisions, all elaborated in \ref{QA12} and \ref{QA3}. For these four questions, we utilized GPT-4-turbo to automatically annotate a total of 1000 multi-turn Q\&A examples based on our proposed workflow framework, with 250 examples for each question. Subsequently, we finetuned this dataset on LLaMA3-8B. The dataset we constructed and the parameter weights after finetuning are available in the GitHub link in abstract.



\section{Experiments}

\begin{table*}[t] \scriptsize
\centering
\setlength{\tabcolsep}{0.2mm}
\renewcommand\arraystretch{1.5}
\begin{tabular}{>{\centering\hspace{0pt}}m{0.073\linewidth}|>{\centering\hspace{0pt}}m{0.048\linewidth}>{\centering\hspace{0pt}}m{0.052\linewidth}>{\centering\hspace{0pt}}m{0.056\linewidth}>{\centering\hspace{0pt}}m{0.048\linewidth}|>{\centering\hspace{0pt}}m{0.054\linewidth}>{\centering\hspace{0pt}}m{0.058\linewidth}|>{\centering\hspace{0pt}}m{0.052\linewidth}>{\centering\hspace{0pt}}m{0.056\linewidth}|>{\centering\hspace{0pt}}m{0.048\linewidth}>{\centering\hspace{0pt}}m{0.052\linewidth}>{\centering\hspace{0pt}}m{0.056\linewidth}>{\centering\hspace{0pt}}m{0.048\linewidth}|>{\centering\hspace{0pt}}m{0.054\linewidth}>{\centering\hspace{0pt}}m{0.058\linewidth}|>{\centering\hspace{0pt}}m{0.052\linewidth}>{\centering\arraybackslash\hspace{0pt}}m{0.056\linewidth}} 
\hline
\multicolumn{9}{>{\centering\hspace{0pt}}m{0.496\linewidth}|}{\textbf{Tencent}}                                                                                                                                                                                                                                                                                                        & \multicolumn{8}{>{\centering\arraybackslash\hspace{0pt}}m{0.424\linewidth}}{\textbf{ChinaMobile}}                                                                                                                                                        \\ 
\hline
\multicolumn{1}{>{\Centering\hspace{0pt}}p{0.033\linewidth}|}{\multirow{2}{0.034\linewidth}{\hspace{0pt}\textbf{Model}}} & \multicolumn{4}{>{\Centering\hspace{0pt}}p{0.204\linewidth}|}{\vcell{\textbf{Statistical}}} & \multicolumn{2}{>{\Centering\hspace{0pt}}p{0.112\linewidth}|}{\vcell{\textbf{Semantic}}} & \multicolumn{2}{>{\Centering\hspace{0pt}}p{0.108\linewidth}|}{\vcell{\textbf{Aggregated}}} & \multicolumn{4}{>{\Centering\hspace{0pt}}p{0.204\linewidth}|}{\vcell{\textbf{Statistical}}} & \multicolumn{2}{>{\Centering\hspace{0pt}}p{0.112\linewidth}|}{\vcell{\textbf{Semantic}}} & \multicolumn{2}{>{\Centering\hspace{0pt}}p{0.108\linewidth}}{\vcell{\textbf{Aggregated}}}  \\[-\rowheight]
\multicolumn{1}{>{\Centering\hspace{0pt}}m{0.073\linewidth}|}{}                                                             & \multicolumn{4}{>{\Centering\hspace{0pt}}m{0.204\linewidth}|}{\printcellbottom}    & \multicolumn{2}{>{\Centering\hspace{0pt}}m{0.112\linewidth}|}{\printcellbottom} & \multicolumn{2}{>{\Centering\hspace{0pt}}m{0.108\linewidth}|}{\printcellbottom}   & \multicolumn{4}{>{\Centering\hspace{0pt}}m{0.204\linewidth}|}{\printcellmiddle}    & \multicolumn{2}{>{\Centering\hspace{0pt}}m{0.112\linewidth}|}{\printcellmiddle} & \multicolumn{2}{>{\Centering\hspace{0pt}}m{0.108\linewidth}}{\printcellmiddle}    \\ 
\cline{2-17}
\multicolumn{1}{>{\Centering\hspace{0pt}}m{0.073\linewidth}|}{}                                                             & radius          & dayloc        & itdNum      & g-rank                       & itdErr       & itdType                                                    & locfreq         & odSim                                                           & radius          & dayloc        & itdNum      & g-rank                       & itdErr       & itdType                                                    & locfreq         & odSim                                                         \\ 
\hline
V-LLM                                                                                                                 & 0.113          & 0.465         & 0.634          & 0.647                       & 0.578          & 0.187                                                        & 0.603          & 6.80                                                        & 0.048          & 0.209          & 0.138          & 0.057                       & 0.642          & 0.198                                                        & 0.436          & 6.59                                                        \\
ST+G                                                                                                                        & 0.029          & 0.163          & \underline{0.076}          & 0.027                       & 0.477          & 0.101                                                        & 0.352          & 6.21                                                        & 0.044          & 0.159          & 0.111           & 0.035                       & 0.572          & 0.124                                                        & 0.401          & 6.34                                                        \\
COT+G                                                                                                                       & 0.028          & 0.093          & 0.080          & 0.018                       & 0.374          & 0.074                                                        & \underline{0.284}          & 5.80                                                        & 0.031          & \underline{0.094}          & \underline{0.083}          & 0.010                       & 0.436          & 0.064                                                        & \underline{0.284}          & 5.76                                                        \\
TOT+G                                                                                                                       & 0.028          & 0.102          & 0.084          & 0.020                       & \underline{0.351}          & 0.108                                                        & 0.286          & 5.83                                                        & 0.033          & 0.095          & 0.084          & \underline{0.009}                       & 0.417          & 0.083                                                        & 0.285          & 5.65                                                        \\ 
\hline
TimeGeo                                                                                                                     & 0.259          & 0.251          & 0.204          & 0.018                       & 0.536          & 0.155                                                        & 0.693          & 7.38                                                        & 0.262        & 0.276          & 0.220          & 0.018                       & 0.544          & 0.155                                                        & 0.639          & 8.40                                                        \\
MoveSim                                                                                                                     & 0.224          & 0.052          & 0.101          & 0.024                       & 0.904          & 0.178                                                        & 0.638          & 5.93                                                        & 0.537          & 0.684          & 0.652         & 0.044                       & 0.879          & 0.278                                                        & 0.454          & 6.59                                                        \\
Volunteer                                                                                                                   & 0.512          & 0.056          & 0.330          & 0.021                       & 0.804          & 0.162                                                        & 0.296          & 6.00                                                        & 0.473         & 0.651          & 0.639          & 0.027                       & 0.863          & 0.220                                                        & 0.436         & 7.00                                                        \\
DiffTraj                                                                                                                    & 0.028          & 0.693          & 0.693         & 0.029                    & 0.597          & 0.080                                                        & 0.287        & 6.38                                                        & \underline{0.025}          & 0.694          & 0.693        & 0.015                      & 0.554         & 0.060                                                   & 0.294          & 5.38                                                        \\ 
Act2Loc                                                                                                                    & \underline{0.021}    & \underline{0.035}          & 0.102         & \underline{0.010}                    & 0.391          & \underline{0.040}                                                        & 0.294        & \underline{5.45}                                                        & 0.028          & 0.143          & 0.158        & \underline{0.009}                      & \underline{0.416}         & \underline{0.048}                                                  & 0.305          & \underline{5.35}                                                        \\ 
\hline
\textbf{CoPB}                                                                                                        & \textbf{0.019} & \textbf{0.029} & \textbf{0.071} & \textbf{0.009}              & \textbf{0.194} & \textbf{0.034}                                               & \textbf{0.282} & \textbf{5.32}                                               & \textbf{0.024} & \textbf{0.066} & \textbf{0.062} & \textbf{0.004}              & \textbf{0.247} & \textbf{0.047}                                               & \textbf{0.267} & \textbf{5.27}                                               \\
\hline
\end{tabular}

\vspace{2mm}
\caption{Performance comparison between our model and baselines on Tencent and ChinaMobile datasets. All models involving LLMs use GPT-4-turbo to conduct experiments. All the metrics are better when smaller. \textbf{Bold} denotes the best results and \underline{underline} denotes the second-best results. Please note that the unit for the column odSim is \textit{E-05}.}
\label{tbl:mainResult}
\end{table*}

\vspace{-4mm}

In this section, we carry out experiments to verify the validity of CoPB workflow. To demonstrate the improvement in capability of generating mobility behaviour by the \textit{Theory of Planned Behaviour}, we compare CoPB with vanilla prompting method, as well as COT and TOT, in terms of the quality of generated data. Several powerful deep generative models are also included in the comparison list.
Additionally, we explore how mechanistic model can lessen the token cost to illustrate the synergistic effect. Finally, we demonstrated that fine-tuned LLaMA3-8B using a self-constructed dataset can further reduce time and monetary costs while maintaining quality.


\subsection{Experimental Setup}

\textbf{Datasets.} Our experiments are conducted on two real-world mobility datasets collected from two data sources, Tencent and ChinaMobile, covering the geographical area of Beijing, China. The basic information of these datasets is presented in Appendix \ref{appendix:dataset}.

\textbf{Baselines.} To rigorously evaluate the improvements our framework has achieved in mobility generation inference, we have established the following inference baselines: \textbf{\textit{V-LLM}}: vanilla LLM. Without any explicit inference mechanism in place, we provided the LLM with nearby POI information and tasked it with autonomously scheduling a day's agenda, including selecting POI destinations to execute these plans.
\textbf{\textit{ST+G}}: \underline{S}imple \underline{t}houghts organization with gravity model. Separating intention generation and POI selection, we first allow the LLM to directly generate a day's intent schedule in one step. Then, we utilize gravity model to map the abstract intentions to specific geographical POIs.
\textbf{\textit{COT+G}}: Intention generation segment employs Chain-of-Thought (COT) for incremental reasoning. Similarly, the gravity model is utilized.
\textbf{\textit{TOT+G}}: TOT for reasoning and gravity model for mapping. Moreover, to objectively evaluate the absolute quality of the model, we also compared it with classic and emerging mobility generation models. These baselines can be categorized into 2 groups, simple mechanistic models: TimeGeo\cite{jiang2016timegeo}, and deep generative models: MoveSim\cite{feng2020learning}, Volunteer\cite{long2023VAE}, DiffTraj\cite{zhu2023difftraj} and Act2Loc\cite{liu2024act2loc}. Detailed information about these baselines is presented in Appendix \ref{appendix:Baselines}.

\textbf{Evaluation Methods.}
In order to make an comprehensive evaluation of the generated mobility behaviour, we evaluate the quality from three dimensions: statistical evaluation, semantic evaluation and aggregation evaluation.
\begin{itemize}[leftmargin=*]
    \item  \textbf{Statistical Evaluation.} For the statistical metrics, we calculate the distance of the distribution of each statistic between the generated data and the real data with Jensen–Shannon divergence(JSD). 
    Specifically, we will quantitatively calculate the JSDs on the following 4 metrics:  \textit{radius}, \textit{dayloc}, \textit{itdNum}, \textit{g-rank}. We refer the readers to Appendix \ref{appendix:Statistical-Metrics} for more details.
    
    \item  \textbf{Semantic Evaluation.} In previous work, assessments have been limited to the statistical level, while here we add evaluation on the semantic dimension by means of two metrics: intention error (\textit{itdErr}) and the distribution of intention categories (\textit{itdType}). The first metric measures the discrepancy between the generated intention sequence for a given persona and the actual intention sequence of this persona's population. The second metric assesses the proportion of time each intention category occupies within the intention sequence. Specific calculation methods and explanations can be found in Appendix \ref{appendix:Semantic-Metrics}.

    \item \textbf{Aggregation Evaluation.} We also evaluate the authenticity of the spatial distribution of the data from an aggregated perspective. Here, we primarily utilized two metrics: \textit{locfreq} and \textit{odSim}. The former metric, \textit{locfreq}, assesses the overall distribution of trajectories on the map by partitioning the map into grid points. The latter metric, odSim, calculated OD (Origin-Destination) matrices based on trajectories, examining the error between the OD matrices of generated data and real data. We refer the readers to Appendix \ref{appendix:Aggregation-Metrics} for more details.

\end{itemize}

\subsection{Improved behaviour Inference Performance.}

Tables \ref{tbl:mainResult} show the full results of our evaluation of the three dimensions on the Tencent and ChinaMobile datasets. As we focus solely on the discrepancies between the proposed generation workflow and existing LLM inference methods in this part, attention can be directed solely to the first four baselines in the table.
As anticipated, the pure LLM version without the use of the gravity model performed the poorest. This suggests that LLM alone cannot efficiently make the leap from mobility intention to specific locations. Neither the accuracy at the intention level nor the statistical metrics were satisfactory. It is worth noting that we provided distance information from the current location to candidate POIs in the inputs, indicating a low sensitivity of LLM to distance and the challenge of aligning with the distribution of real trajectories.
However, once we abstract the mapping of geographical locations separately from mobility behaviours and detach it from intention generation, using a specialized gravity model to handle location mapping as done in \ref{method-gravity}, we observe a significant improvement in generation quality, especially at the numerical statistical level. Furthermore, the intervention of COT notably enhances the quality of generation. However, the introduction of TOT does not bring significant amplification over COT. In fact, some metrics even decline. Upon analyzing the intention results and reasoning steps generated by TOT, we found that TOT's evaluation mechanism naturally inclines the LLM to select more conservative and common intentions, such as work and home, substantially reducing the diversity of trajectory generation. This aligns with the intuition behind TOT's proposition, as it essentially guides the reasoning process in a correct direction. In mathematics problems, a comparative evaluation can lead to a relatively correct direction. While in mobility behaviour generation, defining \textit{correctness} itself poses challenge.

Here we also compare our framework with one classic mechanistic model and several powerful deep generative models to illustrate the advantages of our inference framework. 
As shown in the results in table \ref{tbl:mainResult}, the quality of intention generation of these baselines is far inferior to ours. 
Even Act2Loc, which has a specialized independent transformer for intention generation, cannot achieve accurate and diverse generation. But Act2Loc's performance in numerical and statistical evaluations is very close to ours, demonstrating the effectiveness of the divide-and-conquer approach. The results of other baseline models vary, with DiffTraj performing relatively well, likely due to the powerful spatial distribution modeling capability of diffusion models.

It is worth mentioning that although the original dataset has a huge amount of data, such as the Tencent dataset with 100,000 trajectories, only \textbf{200} pieces of trajectories are used in our model, while the other baselines use 70\% of the whole dataset for training.
Additionally, we conducted ablation experiments to separately examine the effectiveness of each component of the \textit{Theory of Planned Behaviour}, providing more interpretability for the constructed workflow. Please refer to Appendix\ref{appendix:ablation} for detailed content.

\begin{figure*}[t]
    \centering
    \includegraphics[width=0.9\linewidth]{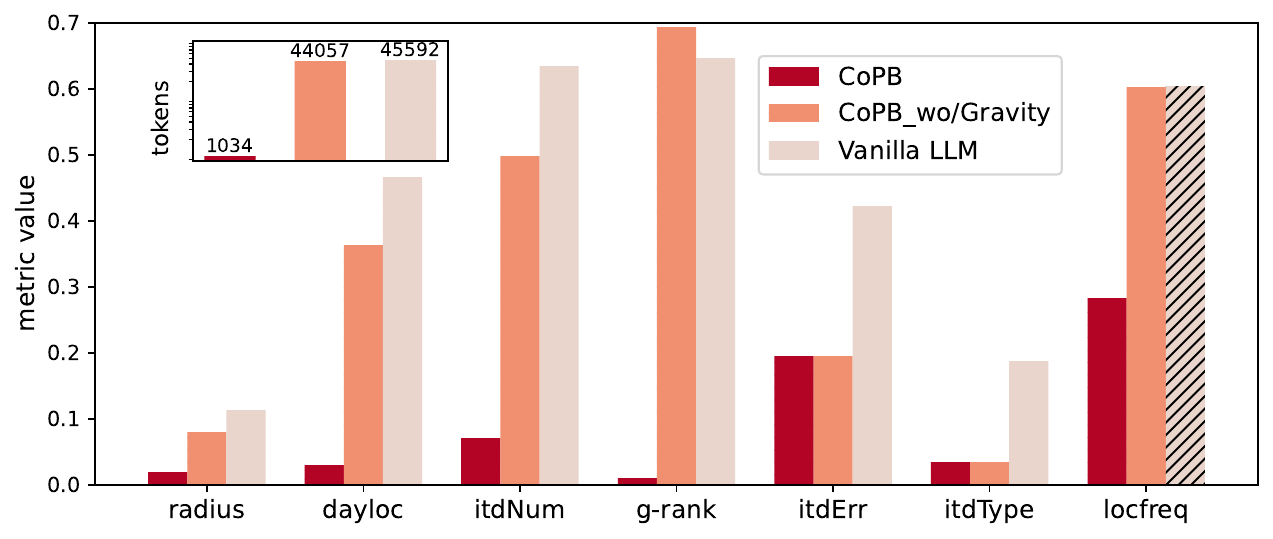}
    \vspace{-2mm}
    \caption{Improvements in saving token cost and the quality of generated results brought by mechanistic model.}
    \label{fig:mechanism}
\end{figure*}

\subsection{The Synergistic Effects and Performance Gains of the Mechanistic Model.}
As illustrated in figure \ref{fig:mechanism}, besides the vanilla LLM version for baseline, we also established an additional baseline where we removed the gravity model from the existing workflow and allowed LLM to select the next POI from the externally provided POI information (including poi names and distance values) on its own by considering preference and distance factors, which is donated as CoPB\textunderscore wo/Gravity in the figure. We visualized the differences in terms of metrics and average token consumption per trajectory. The results reveal that our method not only consumes significantly fewer tokens than the two baselines but also greatly surpasses them in trajectory quality. This is not only because the use of gravity model can significantly reduce the prompts needed for POI inputs but also due to that the intention-time sequence generated by LLM can be mapped into multiple physical trajectories via the gravity model without sacrificing quality or diversity. In the specific experimental process, we mapped each intent sequence to 20 physical trajectories using a gravity model.

\subsection{Performance and Cost Reduction for the Finetue LLaMA3.}
We constructed a mobility behaviour dataset comprising 1000 multi-turn dialogues. Training on a single A100 GPU with LoRA took only 20 minutes for 2 epochs. Then we tested the proposed workflow on several versions of ChatGPT models and compared it with the original LLaMA3-8B model and its fine-tuned version (donated as LLaMA3-8B-F) on Tencent dataset. 
Results in table \ref{tbl:llm-versions} indicate that GPT-4 and GPT-4-turbo performed very similarly, but better than GPT-3.5. The original LLaMA3-8B performed poorly, but after finetuning, it achieved performance similar to GPT-4. 
We also compare the api cost and time consumption of generating 1 thousand pieces of intention sequences using these models. We uniformly leverage one API key for single-threaded API calls for all GPT models and deploy LLaMA3-8B on a local A100 GPU for inference. The token calculation is based on the latest official OpenAI pricing at the time of submission and all experiments are carried out under the same network condition. Results reveal that the finetuned LLaMA3-8B can cut the generation time in half, reduce the cost of the api to zero, thus greatly facilitating the model's large-scale and widespread application downstream..

\begin{table*} [t]
\small
\renewcommand\arraystretch{1.1}
\centering
\setlength{\tabcolsep}{0.1mm}
\begin{tabular}{>{\centering\hspace{0pt}}m{0.183\linewidth}|>{\centering\hspace{0pt}}m{0.067\linewidth}>{\centering\hspace{0pt}}m{0.071\linewidth}>{\centering\hspace{0pt}}m{0.081\linewidth}>{\centering\hspace{0pt}}m{0.071\linewidth}|>{\centering\hspace{0pt}}m{0.062\linewidth}>{\centering\hspace{0pt}}m{0.077\linewidth}|>{\centering\hspace{0pt}}m{0.075\linewidth}>{\centering\hspace{0pt}}m{0.071\linewidth}|>{\centering\hspace{0pt}}m{0.092\linewidth}>{\centering\arraybackslash\hspace{0pt}}m{0.081\linewidth}} 
\hline
\multicolumn{1}{>{\Centering\hspace{0pt}}p{0.153\linewidth}|}{\multirow{2}{0.183\linewidth}{\hspace{0pt}\Centering{}\textbf{Model}}} & \multicolumn{4}{>{\Centering\hspace{0pt}}p{0.29\linewidth}|}{\vcell{\textbf{Statistical}}} & \multicolumn{2}{>{\Centering\hspace{0pt}}p{0.139\linewidth}|}{\vcell{\textbf{Semantic}}} & \multicolumn{2}{>{\Centering\hspace{0pt}}p{0.146\linewidth}|}{\vcell{\textbf{Aggregated}}} & \multicolumn{2}{>{\Centering\hspace{0pt}}p{0.173\linewidth}}{\vcell{\textbf{Cost }}}  \\[-\rowheight]
\multicolumn{1}{>{\Centering\hspace{0pt}}m{0.183\linewidth}|}{}                                                                          & \multicolumn{4}{>{\Centering\hspace{0pt}}m{0.29\linewidth}|}{\printcellbottom}             & \multicolumn{2}{>{\Centering\hspace{0pt}}m{0.139\linewidth}|}{\printcellbottom}          & \multicolumn{2}{>{\Centering\hspace{0pt}}m{0.146\linewidth}|}{\printcellbottom}            & \multicolumn{2}{>{\Centering\hspace{0pt}}m{0.173\linewidth}}{\printcellmiddle}        \\ 
\cline{2-11}
\multicolumn{1}{>{\Centering\hspace{0pt}}m{0.183\linewidth}|}{}                                                                          & radius & dayloc & itdNum & g-rank                                                          & itdErr & itdType                                                                         & locfreq & odSim                                                                            & time-cost & api-cost                                                                  \\ 
\hline
GPT-3.5-turbo                                                                                                                            & 0.021  & 0.049  & 0.080  & 0.015                                                           & 0.233  & 0.038                                                                           & 0.271   & 5.45                                                                             & 170       & 413.6                                                                     \\
GPT-4                                                                                                                                    & 0.020  & 0.030  & 0.073  & 0.011                                                           & 0.196  & 0.035                                                                           & 0.282   & 5.50                                                                             & 263       & 930.6                                                                     \\
GPT-4-turbo                                                                                                                              & 0.019  & 0.029  & 0.071  & 0.009                                                           & 0.194  & 0.034                                                                           & 0.282   & 5.32                                                                             & 123       & 20.68                                                                     \\ 
\hline
LLaMA3-8B                                                                                                                                & 0.022  & 0.038  & 0.074  & 0.013                                                           & 0.315  & 0.065                                                                           & 0.298   & 5.68                                                                             & 92        & 0                                                                         \\
LLaMA3-8B-F                                                                 & 0.018  & 0.029  & 0.073  & 0.010                                                           & 0.196  & 0.036                                                                           & 0.284   & 5.39                                                                             & 92        & 0                                                                         \\
\hline
\end{tabular}
\caption{Performance of using different versions of LLMs in our mobility behavior generation framework. Unit of odSim: E-05. Unit of time-cost is minutes. Unit of api-cost is USD.}
\label{tbl:llm-versions}
\end{table*}


\section{Conclusion.}
Within this paper, we have meticulously fashioned a behavioural generation framework rooted in the \textit{Theory of Planned Behaviour}, thereby markedly augmenting the capabilities of LLMs in emulating human mobility decision-making processes. Additionally, we delved into the symbiotic interplay between LLMs and mechanistic models to surmount the obstacle of acquiring massive environmental information. Leveraging this sophisticated workflow, we curated a dataset for mobility behaviour generation employing GPT-4-turbo, subsequently finetuning the LLaMA3-8B model on it. Remarkably, this endeavor yielded performance akin to the GPT-4 series models, thereby significantly mitigating the cost associated with employing LLMs for mobility behaviour generation and elevating the feasibility of downstream large-scale simulation.

\clearpage

\bibliography{0.neurips_2024}

\clearpage

\section*{Appendix}

\appendix

\section{Limitations and Future Directions.}

There is huge potential for refinement in the usage of geographical environment information. Sometimes we are not only concerned with the starting and ending locations of a movement but also with the journey in between, such as which modes of transportation were used and which routes were taken. Therefore, introducing more detailed road network data or considering different transportation modes could help generate more precise and realistic mobility trajectories. Additionally, the process of selecting POIs is not only related to distance but is also significantly influenced by personal preferences, past visitation tendencies, and the current foot traffic at the POIs. How to comprehensively consider these factors remains an area to be explored. We are also quite interested in the impact of social interactions on mobility behavior. As social beings, our movements are undoubtedly influenced by the behaviors of those around us. The role of such dynamic mechanisms in large-scale mobility behavior remains unexplored.

\section{Comparison of our CoPB with COT and TOT workflows.}


Here we compare the proposed CoPB framework with the COT and TOT reasoning approaches. COT primarily uses multi-order Markov-based sequential inference, while TOT builds on COT by adding tree-structured branching, evaluation, and selection. CoPB, on the other hand, embeds the \textit{Theory of Planned Behaviour} thinking mode into sequential execution. By introducing additional Q\&A processes for attitude, subjective norms, and perceived behavior control as intermediate steps in the inference, CoPB significantly enhances the capability of behaviour generation.
\begin{figure*}[h]
    \centering
    \includegraphics[width=0.65\linewidth]{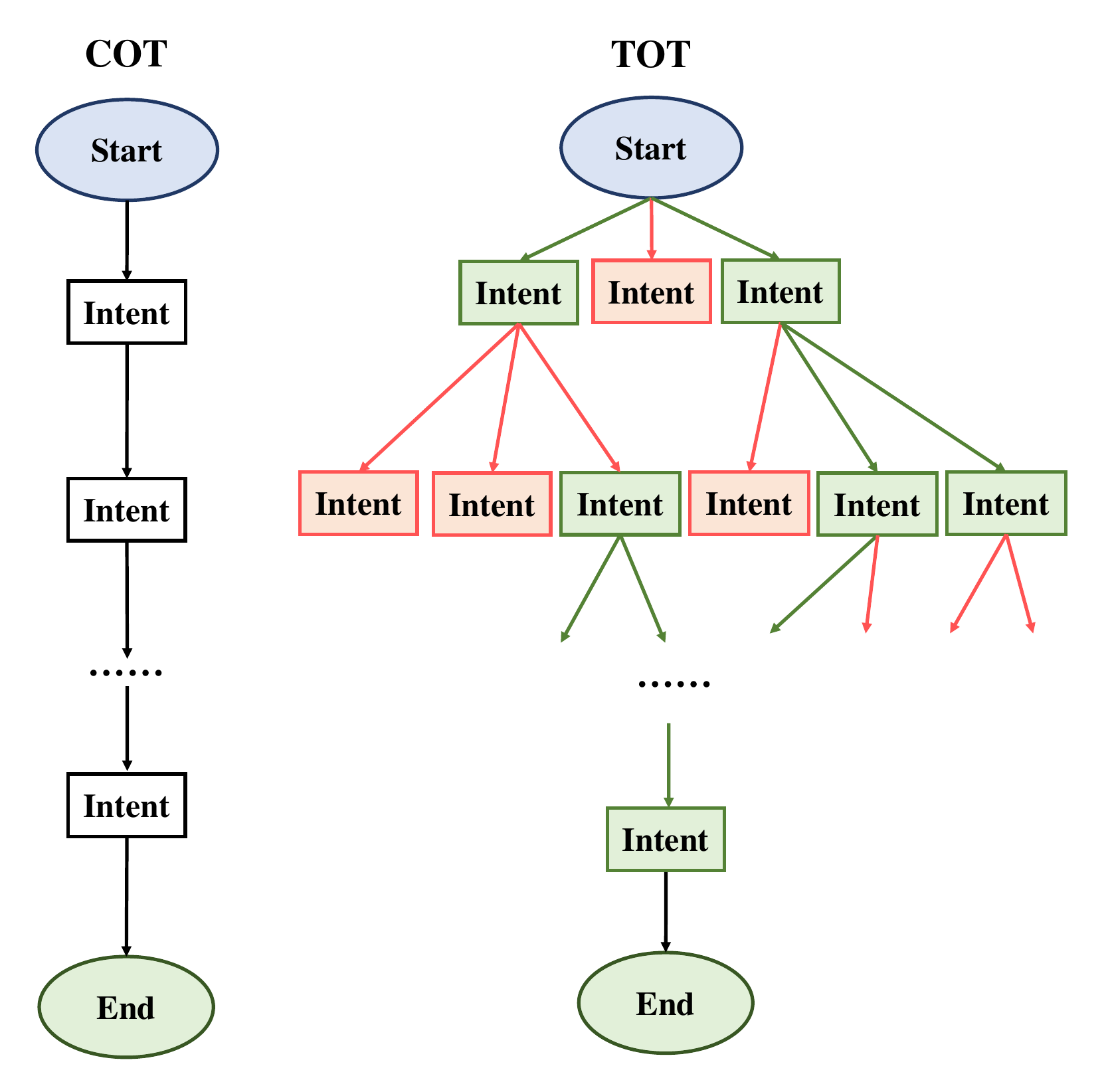}
    \vspace{-3mm}
    \caption{Workflows of COT and TOT.}
    \label{fig:COT-CoPB}
\end{figure*}


\section{Details of the Experimental Setup}

\subsection{Dataset} \label{appendix:dataset}

\textbf{Tencent Dataset}. This dataset is collected from a social network platform and records users’ mobility trajectories. Additionally, users’ profiles, such as income level, gender, occupation, education level and age, are collected through digital surveys. 
We randomly select 200 7-day trajectories from the whole dataset to fine-calibrate the POI matching and manually annotate specific intention types. All the experiments on CoPB use only these 200 trajectories, but for deep generative models, they used 70\% of the entire dataset for training.

\textbf{Mobile Dataset}. This dataset is collected from a local mobile operator and records the time and location of users’ connections to nearby cellular base stations. Similarly, the user profiles are collected through surveys. For each record in this dataset, we match the location of the visit to the nearest POI based on latitude and longitude coordinates, and then characterize the intention of this visit by the category attributes of the POI. We also use only 200 pieces of data to carry out experiments on CoPB. And other deep generative models still used 70\% of the data for training.

The basic information about the trajectory data for both datasets is shown in the following table. The sampling time for both datasets is unfixed.
\begin{table}[H]
\centering

\begin{tabular}{|c|c|c|} 
\hline
\textbf{Datasets}   & \textbf{Tencent}                    & \textbf{ChinaMobile}            \\ 
\hline
Duration            & October 1, 2019 - December 31, 2019 & July 1, 2017 - August 31, 2017  \\ 
\hline
City                & Beijing                             & Beijing                         \\ 
\hline
\#Users             & 100000                              & 1246                            \\ 
\hline
\#Trajectory Points & 297363263                           & 4163651                       \\
\hline
\end{tabular}

\vspace{1mm}
\caption{Basic information about the trajectories of the two datasets}\label{tbl:datasetIntro}
\end{table}
\vspace{-5mm}

From the micro level, there is a close relationship between people's identity information and people's mobility behaviour. Income level, gender, education level, age, job, etc. all affect people's travel patterns. For example, the mobility pattern of an IT professional is likely to be a simple "point-to-point" commute between home and the office while a delivery boy's day may involve commuting between numerous locations in the city. In our experiment on Tencent dataset, the character profile includes 4 kinds of attributes: gender, occupation, education level, and income level. In the experiments on ChinaMobile dataset, the profile includes 3 kinds of attributes: gender, education level, and income level. 
Additionally, due to privacy risks, we did not directly utilize real profile information in the experiment. Instead, based on the locations of homes in trajectories, we statistically derived the demographic distribution of each area. During generation, we initially sampled an area where a home is located and then sampled a specific profile from the demographic distribution of that area.
The distribution of character attributes for both datasets is shown in the following tables.

\begin{table}[H]
\centering

\begin{tabular}{|c|c|} 
\hline
\textbf{User Profile} & \textbf{Category}                                                                                                                                                                                                                                           \\ 
\hline
Income                & \begin{tabular}[c]{@{}c@{}}Low (11.44\%), Slightly Low (18.12\%), \\Medium (28.67\%),Slightly High (18.19\%),~\\High (3.36\%), Uncertain (20.22\%)\end{tabular}                                                                                                \\ 
\hline
Gender                & Male (54.67\%), Female (44.68\%), Uncertain (0.65\%)                                                                                                                                                                                                         \\ 
\hline
Education             & \begin{tabular}[c]{@{}c@{}}Bachelor's degree (25.90\%), High school diploma (21.70\%), \\Master's degree (9.94\%), Elementary school diploma (13.30\%), \\Junior high school diploma (13.05\%), Associate degree (8.02\%),\\Doctoral degree (5.27\%)\end{tabular}  \\ 
\hline
Age                   & \begin{tabular}[c]{@{}c@{}}0-30 (44.50\%), 30-40 (30.40\%), \\40-60 (23.55\%), 60-99 (1.55\%)\end{tabular}                                                                                                                                                  \\ 
\hline
Job(top8)             & \begin{tabular}[c]{@{}c@{}}IT Engineer (11.57\%), Online Sales (8.37\%), \\Training Instructor (7.34\%), Investment (7.31\%),\\Finance (6.97\%), Copywriting (6.91\%), \\Front Desk Reception (6.15\%), Technical Worker (6.05\%)\end{tabular}                      \\
\hline
\end{tabular}

\vspace{1mm}
\caption{Distribution of profiles in Tencent dataset}\label{tbl:tencentProfiles}
\end{table}
\vspace{-5mm}
\begin{table}[H]
\centering

\begin{tabular}{|c|c|} 
\hline
\textbf{User Profile} & \textbf{Category}                                                                                                                                           \\ 
\hline
Income                & \begin{tabular}[c]{@{}c@{}}Low (23.27\%), Relatively Low (20.30\%)\\Medium (36.44\%), Relatively High (15.97\%)\\High (4.01\%)\end{tabular}                      \\ 
\hline
Gender                & Male (63.56\%), Female (36.44\%)                                                                                                                              \\ 
\hline
Education             & \begin{tabular}[c]{@{}c@{}}Bachelor degree (58.43\%), High school degree (21.03\%)\\Master's degree (11.32\%), Junior high school degree (9.23\%)\end{tabular}  \\ 
\hline
Age                   & \begin{tabular}[c]{@{}c@{}}0-30 (22.71\%), 30-40 (28.01\%), \\40-60 (40.85\%), 60-99 (8.43\%)\end{tabular}                                                  \\
\hline
\end{tabular}

\vspace{1mm}
\caption{Distribution of profiles in ChinaMobile dataset}
\end{table}\label{tbl:mobileProfiles}
\vspace{-5mm}

\subsection{Baseline}
\label{appendix:Baselines}

\begin{itemize}[leftmargin=*]
    \item \textbf{TimeGeo}. This work models human mobility behaviour as a chain of probabilistic choices. It defines the weekly home-based tour number, dwell rate and burst rate to model the temporal choices and employs EPR to model the spatial choices.

    \item \textbf{MoveSim}. This model utilizes a generative adversarial framework, wherein the discriminator is continually reinforced to supervise the generator in producing high-quality trajectory data. To introduce prior knowledge about urban spatial structures, an attention-based region network is designed within the generator. Furthermore, leveraging the spatial continuity and temporal periodicity migration patterns, both the generator and discriminator are pretrained to accelerate the learning process.

    \item \textbf{Volunteer}. This work uses two vae for joint modeling, the first user VAE is used to learn the distribution of an individual's residence and workplace, and the second VAE decouples travel time and dwell time to accurately model an individual's movement trajectory.
    
    \item \textbf{DiffTraj}. This work proposes a spatio-temporal diffusion probabilistic model for trajectory generation, the core idea of which is to reconstruct and synthesize geographic trajectories from white noise through a reverse trajectory denoising process. UNet is used to embed conditional information and accurately estimate the noise level in the inverse process.

    \item \textbf{Act2Loc}. This model integrates deep learning models with mechanistic models. It uses a transformer to continue writing new trajectories based on real activity sequences, while also employing the \textit{Universal Opportunity (UO)} model to select geographic grid points based on population distribution. Despite leveraging a divide-and-conquer approach, it still adheres to the paradigm of fitting before sampling in deep generative models. The use of prediction with transformers for generation also limits the model's freedom in generating. 
    
\end{itemize}

\subsection{Statistical Evaluation Metrics} \label{appendix:Statistical-Metrics}

We use JSD (Jensen-Shannon Divergence) to calculate the distance between the distributions of metrics for generated data and real data, which is defined as $JSD(p;q)=h((p+q)/2)-(h(p)+h(q))/2$, where $h$ represents the Shannon information, $p$ and $q$ are two distributions. The lower the JSD, the closer the two distributions are, and the better the quality of generated data is. Detailed information about the four statistics is shown below:

\begin{itemize} [leftmargin=*]
    \item \textbf{radius}. radius of gyration, which represents the spatial range of the user's daily activities.
    \item \textbf{dayloc}. daily visited locations, which is calculated as number of different locations visited per day for each of the user.
    \item \textbf{itdNum}. number of intentions per day, which differs from dayloc in that there is no de-duplication of locations.
    \item \textbf{g-rank}. number of visits to different locations, which is calculated as the top-100 visiting frequency among all the locations.
\end{itemize}

\subsection{Semantic Evaluation Metrics} \label{appendix:Semantic-Metrics}
One preparatory work before the semantic evaluation is needed: matching intention categories for trajectory records. For the real data, we get the intention category information of each location visit record by manual labeling or POI matching. For the results we generate, they naturally contain the category distinction. For the baseline model without the concept of intention category, after we identify the stay at home and working from the trajectories, other intentions are randomly assigned.\par
Then, for the metric \textbf{itdErr}, we need to calculate the similarity by direct comparison between the generated intention sequences and the real ones. Since real data inevitably possesses uncertainty and randomness, we aggregate 7 days of a person's trajectories into 1 day by majority voting on the same time slice. Of course, we only focus on the level of intention type when aggregating. For example, if 5 of these 7 days are eating at 12:00, and 2 days are working at 12:00, then the intention of 12:00 in the aggregated result is eating. With the help of aggregation, the uncertainty of the trajectory is eliminated and a more stable and representative real trajectory of the specified profile is obtained. Then the generated results are compared with the aggregated trajectory on a time-slice-by-time-slice basis. The length of the time slices is standardized to half an hour. For the metric \textbf{itdType}, we calculate the time proportion of each category of intentions in the trajectories weighted by durations, and calculate the JSD of the proportion vectors.

\subsection{Aggregation Evaluation Metrics} \label{appendix:Aggregation-Metrics}
To calculate the metric \textbf{locfreq}, we need to partition the map uniformly into grids, count the frequency of trajectories visiting each grid, and calculate JSD between the frequency vectors. To get \textbf{odSim}, we count the movement of individuals between grids, calculates the OD transfer probability matrix, and computes the MSE values between the matrices. In addition to the numerical measurement, we also visualize the grid-visiting frequency difference between the generated data and real data using heatmaps, such that we can intuitively judge the quality of the generated data through the visual difference between our model and baselines.  

\section{Additional Experiments}

\subsection{\textbf{Visualization of Spatial Errors in Generated Trajectories.}}

\begin{figure*}[h]
    \centering
    \includegraphics[width=1\linewidth]{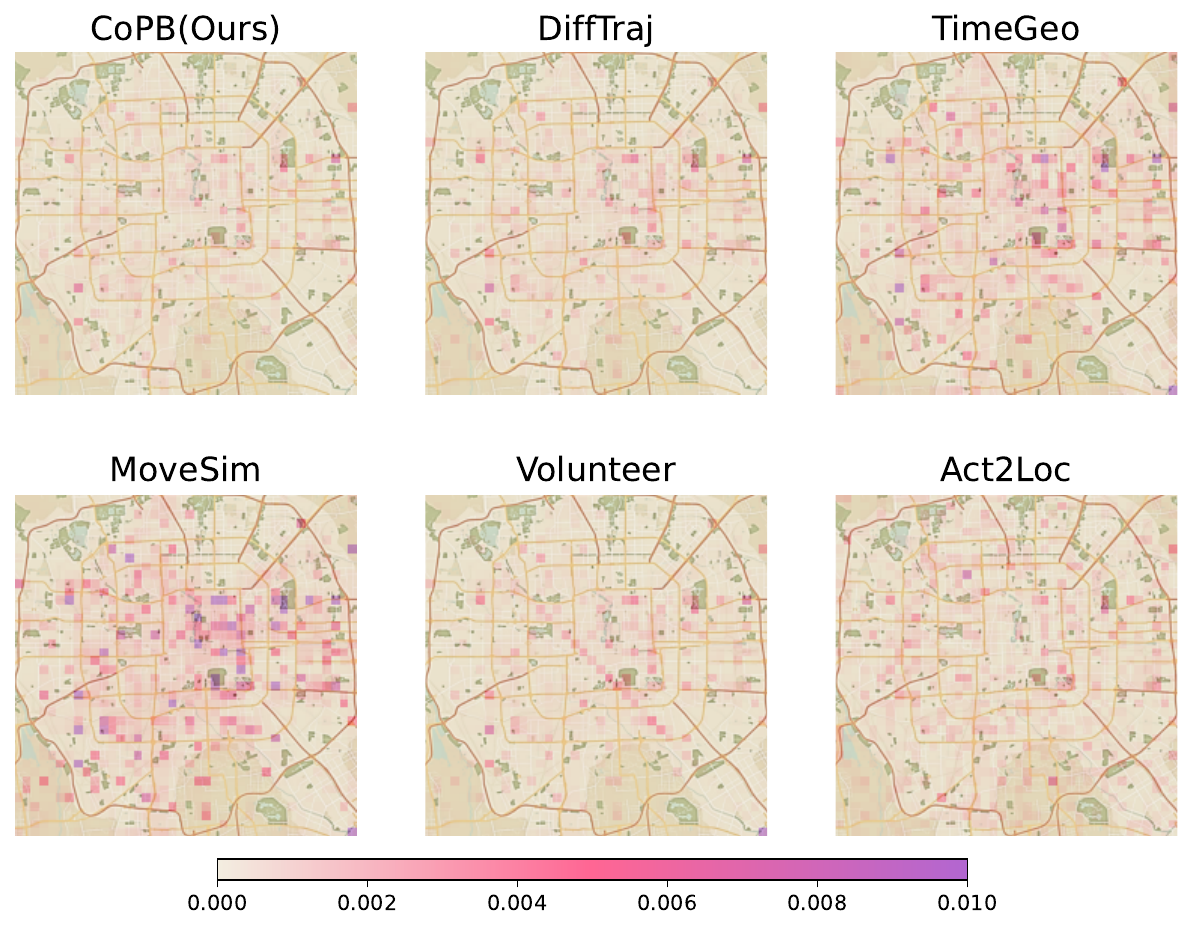}
    \vspace{-3mm}
    \caption{Spatial Errors in Generated Trajectories.}
    \label{fig:spatialerror}
\end{figure*}

In addition to the computation of metrics in three dimensions, we also use heatmaps to visualize the grid-visiting frequency difference between the generated data and real data. We restrict the visualization to a square area connected to the fifth ring road of Beijing and compare the results of our model with mechanistic model or deep learning baselines. 
As shown in Figure \ref{fig:spatialerror}, the color depth of each grid point represents the difference in the visiting frequency compared to the real data. The closer to purple, the larger the error.
The trajectories generated by our model are the closest to the real trajectories as there are very few purple or red grid color blocks on our graph. The results of Act2Loc are visually similar to ours, except that some local areas have slightly darker colors. After all, we are all using mechanistic models as a tool for geo-mapping, and in this context, only the gap in quality of generation at the intention level is left. Results of DiffTraj and Volunteer are acceptable, indicating that Diffusion and VAE have accurate spatial fitting capabilities. But the images of MoveSim and TimeGeo show overall darker colors, indicating that their errors are larger.

\subsection{\textbf{Ablation Study.}} \label{appendix:ablation}
\begin{table*}[h] \small
\centering
\begin{tabular}{c|cccc|cc|cc} 
\hline
\multirow{2}{*}{\textbf{Model}} & \multicolumn{4}{c|}{\vcell{\textbf{Statistical}}}                 & \multicolumn{2}{c|}{\vcell{\textbf{Semantic}}} & \multicolumn{2}{c}{\vcell{\textbf{Aggregated}}}  \\[-\rowheight]
                                & \multicolumn{4}{c|}{\printcellbottom}                             & \multicolumn{2}{c|}{\printcellbottom}          & \multicolumn{2}{c}{\printcellbottom}             \\ 
\cline{2-9}
                                & radius         & dayloc         & itdNum         & g-rank         & itdErr         & itdType                       & locfreq        & odSim                           \\ 
\hline
wo\_A                           & 0.028          & 0.151          & 0.129          & 0.013          & 0.384          & 0.086                         & 0.303          & 5.79                            \\
wo\_SN                          & 0.021          & 0.030          & 0.072          & 0.011          & 0.266          & 0.047                         & 0.308          & 5.38                            \\ 

wo\_PBC                         & 0.022          & 0.037          & 0.079          & 0.015          & 0.289          & 0.051                         & 0.297          & 5.40                            \\

\hline
\textbf{CoPB}                    & \textbf{0.019} & \textbf{0.029} & \textbf{0.071} & \textbf{0.009} & \textbf{0.194} & \textbf{0.034}                & \textbf{0.282} & \textbf{5.32}                   \\
\hline
\end{tabular}

\caption{Abalation study regarding the impact of the three TPB (Theory of Planned Behavior) inference components on the final generation results. Experiment is conducted on the Tencent dataset.}
\label{tbl:ablation}
\end{table*}
To investigate how the three components of the \textit{Theory of Planned Behaviour} separately impact the final inference results and trajectory quality, we conducted ablation experiments by removing one component at a time and evaluating the generated movement behaviours. Attitude, subjective norms, and perceived behavioural control are abbreviated as \textbf{A}, \textbf{SN}, and \textbf{PBC}, respectively.
All experiments still used the gravity model to map intentions to geographic POIs. Results are shown in Table \ref{tbl:ablation}, we found that the removal of attitude had the greatest impact on the results. Further checks revealed that without the attitude closely associated with the profile, the LLM-generated intention sequences tended to homogenize, with time spent at home and work showing convergence. This not only affected the accuracy of intention generation for the specified population but also influenced the statistical properties of the spatial trajectories due to the density of the intention sequences. The impacts of removing perceived behavioural control and subjective norms were similar, both resulting in some performance degradation. However, the impact was more severe when subjective norms were removed. This is because perceived behavioural control relies on the personal anchor points or routines provided by subjective norms to assess the perceived likelihood of the next intention. Without subjective norms, the performance gain from perceived behavioural control also diminishes.

\subsection{\textbf{Few-shot Performance Exploration}}
\begin{figure}[h]
    \centering
    \includegraphics[width=0.95\linewidth]{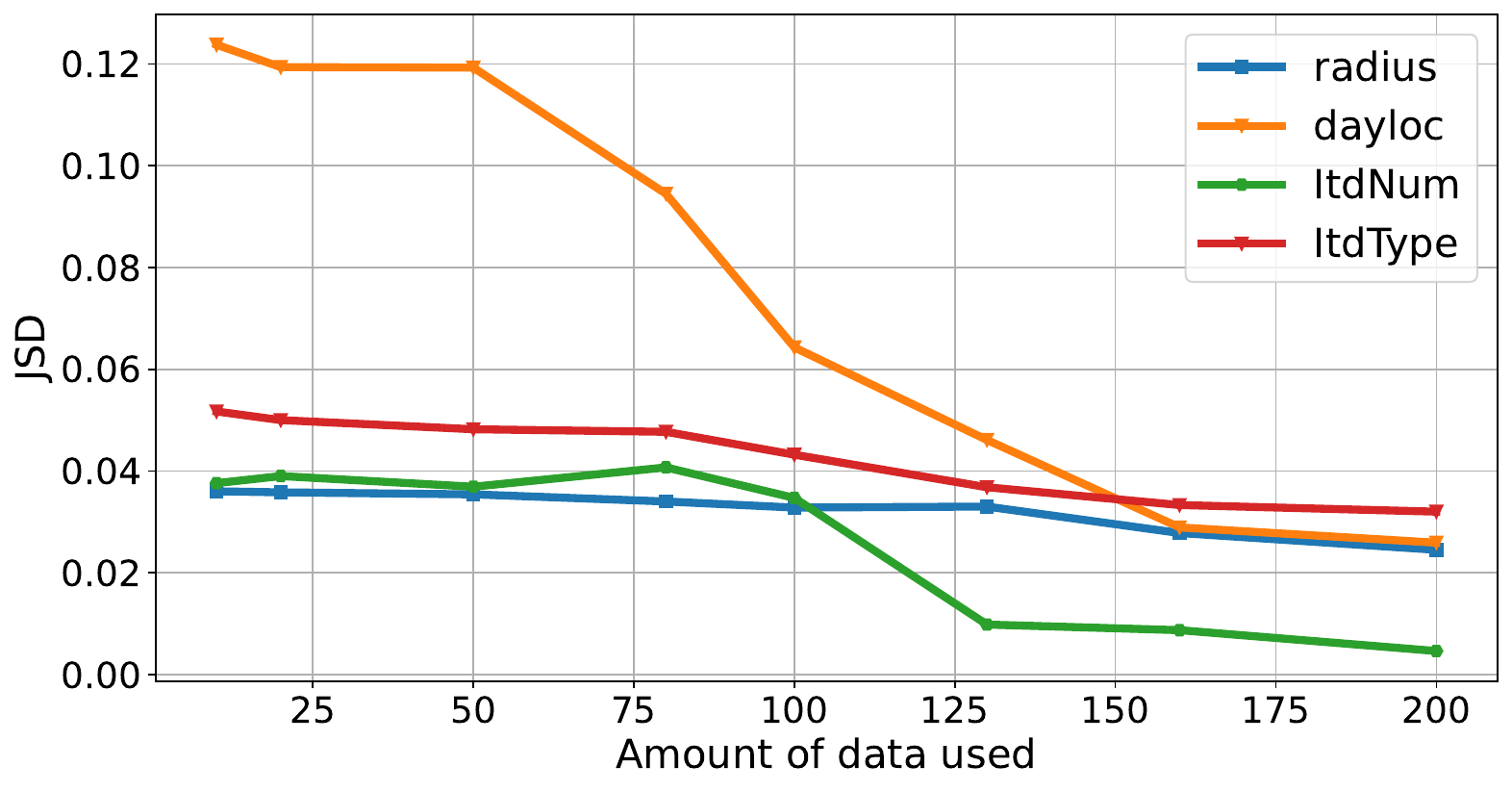}
    \caption{Impact of different training sample sizes on the performance of CoPB.}
    \label{fig:amount}
\end{figure}

In contrast to previous deep generative models that required large amounts of data to fit distributions, our model requires only a very small amount of data. They are mainly used to count the distribution of the number of intentions versus time and to fit the exponent of distance decay in gravity models. In our experiment, only 200 trajectories of 7 days were used. Here, we explore the effect of different amounts of data on the experiment and the results are shown in Figure \ref{fig:amount}.
As can be seen from the figure, the results for most of the metrics are getting better as the amount of data used increases. Statistical metrics like itdType and radius are less affected by the amount of data. The former is generated autonomously by the LLM and does not rely on real data distribution, while the latter does not require much real data to be fitted. However, metrics like dayloc and itdNum are highly affected, with a clear trend of optimization as the amount of data rises, suggesting that the learning of these movement patterns requires a higher amount of data. However, the convergence trend shown at the end of all the curves indicates that our model does not require a large amount of data to exhibit satisfactory generation quality, which is beyond the reach of deep generative models.

\subsection{\textbf{Application of Data Augmentation}}
In this experiment, we will explore the utility of the generated data in real-world applications. Here, we consider a realistic scenario where a mobile service operator's trajectory data cannot be shared with other companies due to issues such as privacy and permissions, and therefore cannot be applied to downstream real-world application tasks. Given the lack of data volume, our model can generate high-quality data to augment the sparse amount of data in applications.

For the selection of specific applications, we choose the mobility prediction task which is to predict the future trajectory points based on the historical trajectory. The prediction task requires accurate mining and modelling of motion patterns in historical data, and is therefore well suited for checking whether the generated data has reasonable patterns and distribution characteristics that are close to the real ones.

\begin{figure}[h]
    \centering
    \subfigcapskip=-3pt 

    \subfigure[Location Prediction Accuracy]{
        \label{fig:prediction1}
        \includegraphics[width=0.46\linewidth]{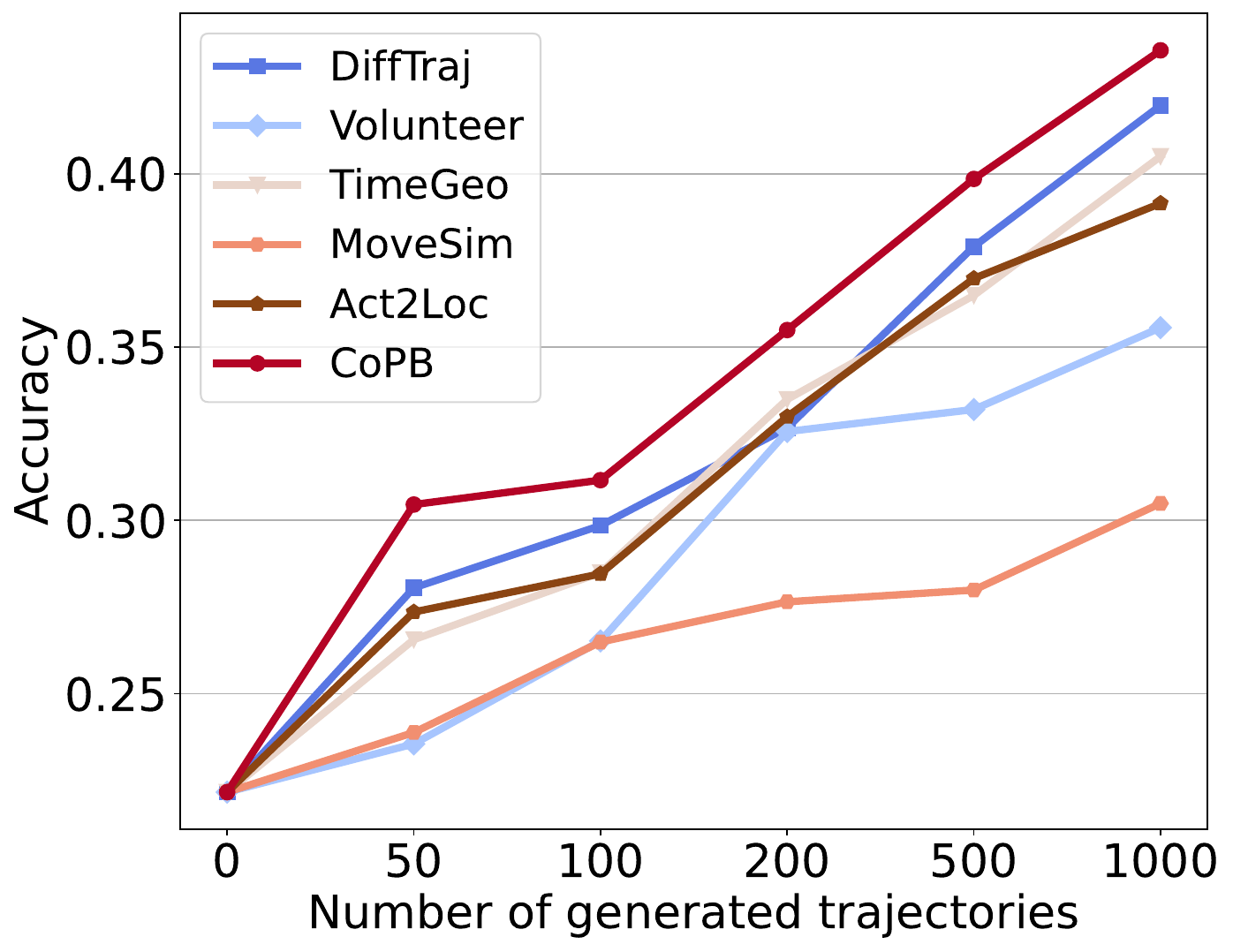}
        }
    \quad
    \subfigure[Intention Prediction Accuracy]{
        \label{fig:prediction2}
        \includegraphics[width=0.46\linewidth]{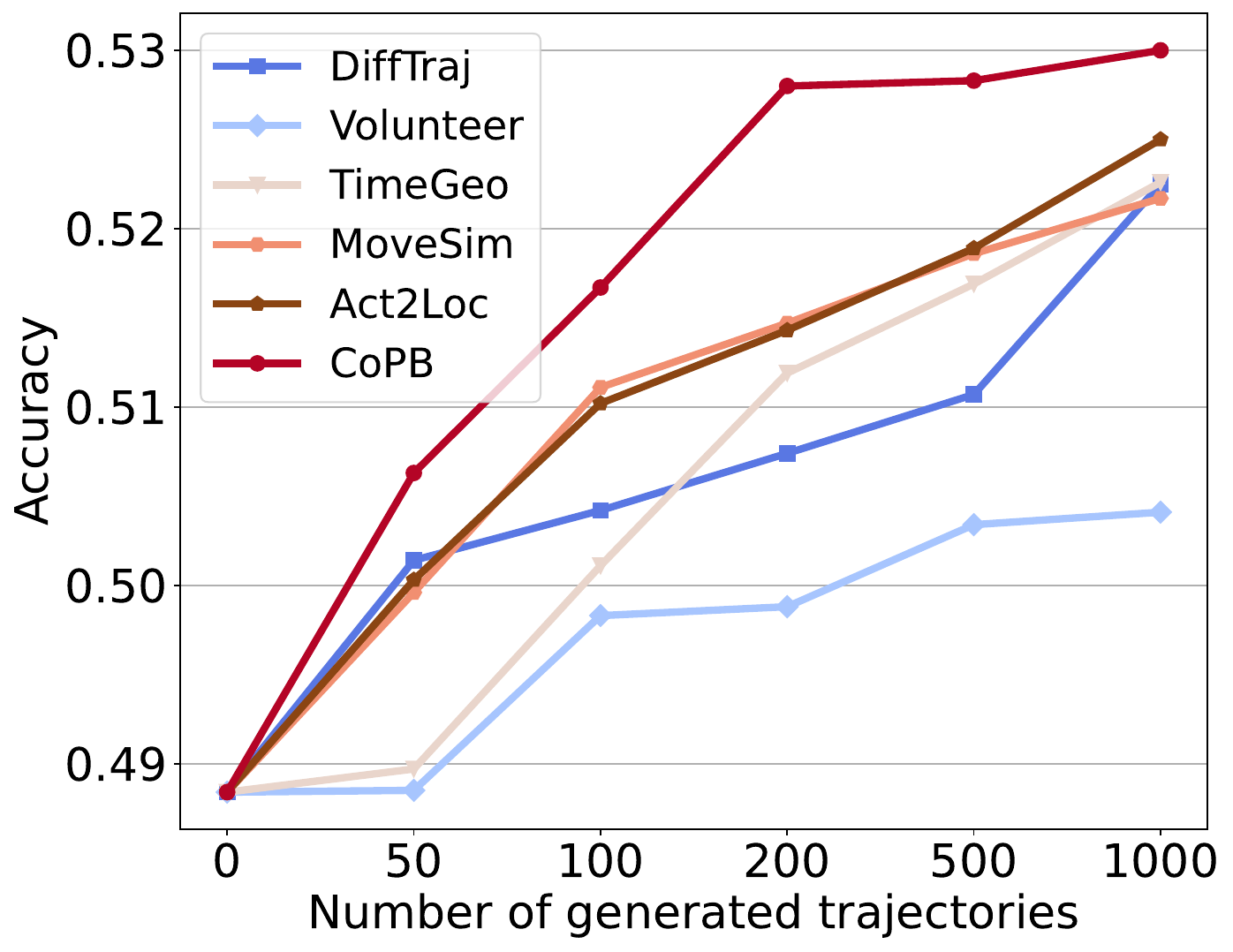}}

    \caption{Performance boost in downstream mobility prediction task when different sizes of generated mobility data is provided as data augmentation.
    }
    \label{fig:prediction}
\end{figure}

We conduct this experiment on the Tencent dataset, dividing the task into geospatial prediction and semantic-level activity type prediction, both using accuracy to measure the prediction performance. The results of the experiment are shown in Figure \ref{fig:prediction}. We fixed the amount of real data at 100, then added varying amounts of generated data to examine the gain in predictive performance. As can be seen from the figure, the generated data from our model all have better enhancements compared to the data generated from baseline models, which is particularly evident in the intention type prediction.

\subsection{\textbf{Privacy Risk}} \label{appendix:privacy}
As mentioned in the introduction, privacy risks are one of the reasons hindering the current large-scale sharing and application of mobile data. Therefore, here we will discuss some experiments on the privacy issues of the proposed framework in generating data. Here we carry out two experiments on privacy.

\begin{itemize} [leftmargin=*]

    \item \textbf{Uniqueness Testing}. We select sequences from generated data and compare them with real ones from trainset in a slice-by-slice way. The overlapping ratio is defined as the ratio of the number of identical slices to the total length. Results tested on the Tencent dataset indicate that, the average overlapping rate between our generated intentions and real ones is only 21\%. Due to the relatively fixed time periods for both work and sleep intentions, such as sleep typically occurring between 11 p.m. and 8 a.m. the next day, and work happening between 9 a.m. and 6 p.m., the value of 21\% can not considered high. If considering location accuracy, the average overlapping rate is only 2‱, and the average maximum overlapping rate(real trajectory most similar to the generated one) is only 5.9‰, demonstrating that our framework is indeed able to generate brand-new and unique trajectories rather than simply copying.
    
    \item \textbf{Membership Inference Attacks}: Membership Inference Attacks (MIAs) arose from concerns about data privacy in machine learning. These attacks determine if a specific data point was part of a model's training set by analyzing the model's outputs. MIAs help evaluate a model's privacy protection: a high success rate indicates potential privacy risks, while a low success rate suggests stronger privacy safeguards. Therefore, MIAs are a very powerful method for assessing and improving the privacy robustness of machine learning models. Our experimental setup follows the configurations in \cite{yuan2024generating} and \cite{lin2020using}. To control for the impact of classification methods, we used two classic and commonly employed classifiers: Random Forest and Logistic Regression, to carry out the attacks. The positive samples are individuals from the 200 training trajectories, while the negative samples are from the remaining dataset. The feature used is the overlap ratios from multiple runs. The metric for evaluation is the success rate, defined as the percentage of successful attempts in determining whether a sample is part of the training dataset.
    The attack success rate of Logistic Regression on the Tencent dataset is 0.512, and on the mobility dataset, it is 0.503. The attack success rate of Random Forest on the Tencent dataset is 0.521, and on the mobility dataset, it is 0.501. The results mean both attackers can hardly infer whether data are in the training set. Therefore, our mobility generation framework poses almost no privacy concerns. This is to be expected, as our approach does not directly input large volumes of trajectory data to fit the model like deep generative models. Instead, we use only 200 trajectories to indirectly derive distributional information. Moreover, the intention sequences in the in-context examples do not even contain specific geographic locations, further mitigating privacy risks.
    
\end{itemize}

\subsection{\textbf{Diversity Evaluation}}

\begin{figure}[htbp]
    \centering
    \subfigure{
        \includegraphics[width=0.31\textwidth]{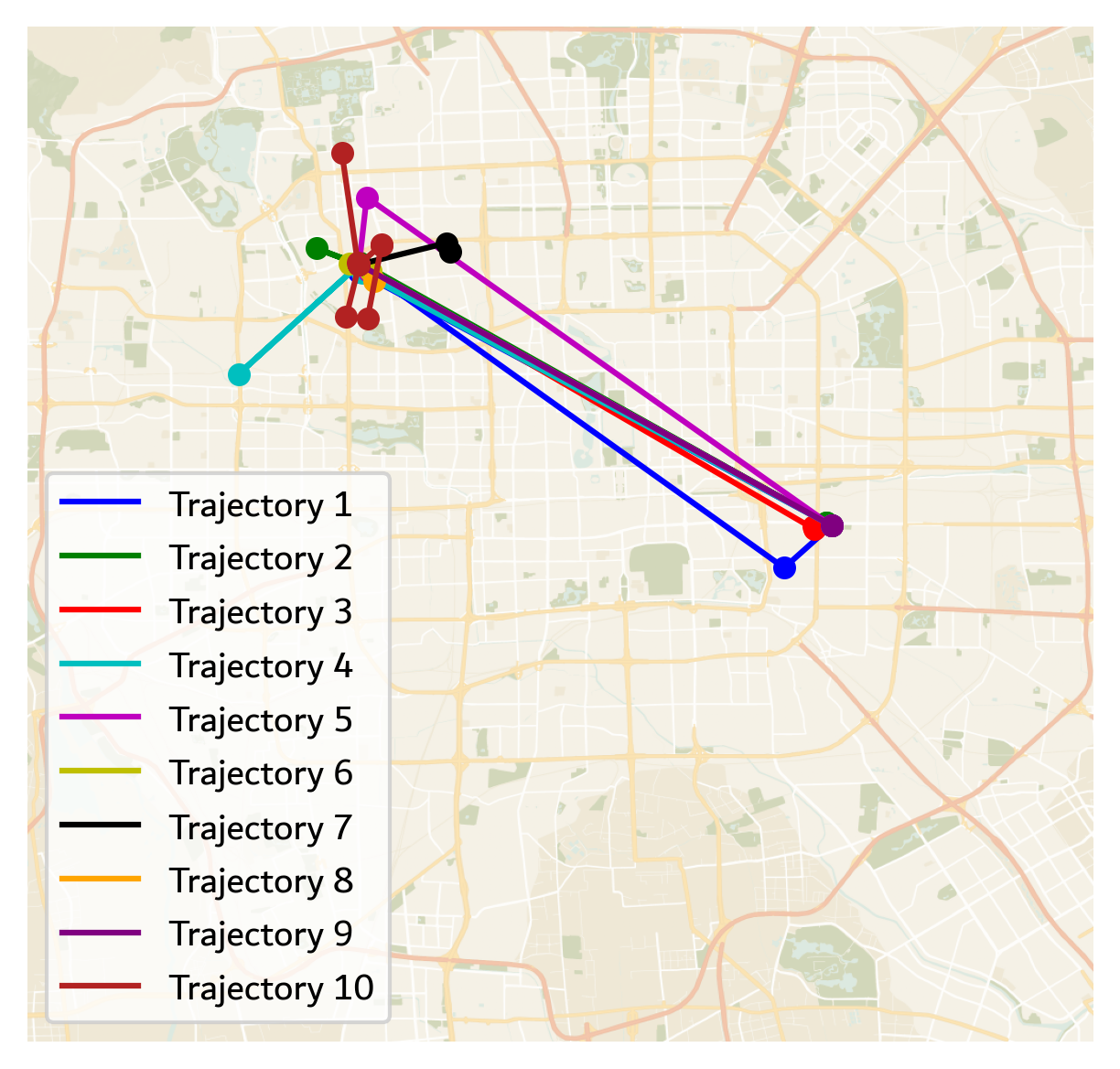}
        \label{fig:sub1}
    }
    \hfill
    \subfigure{
        \includegraphics[width=0.31\textwidth]{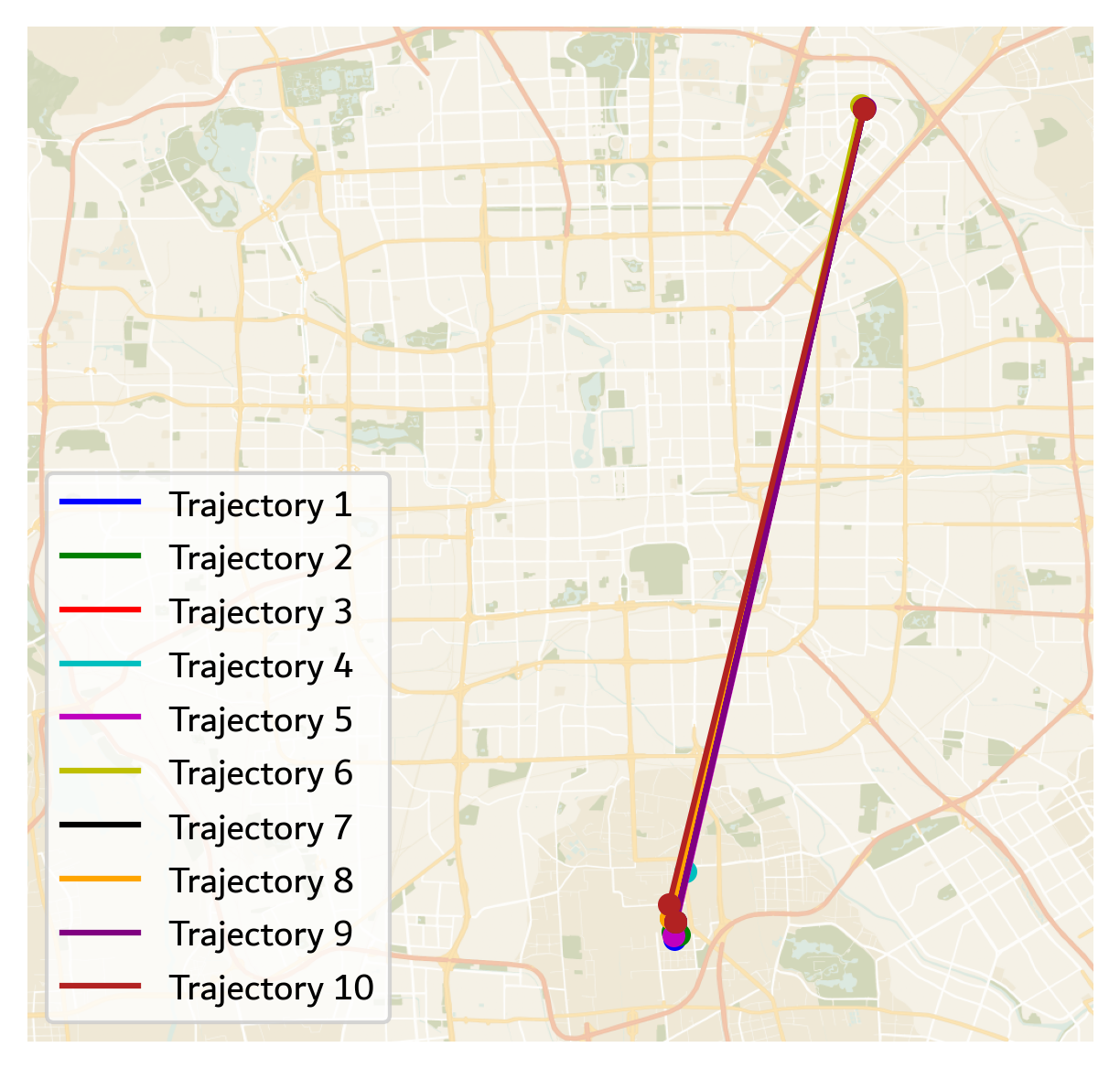}
        \label{fig:sub2}
    }
    \hfill
    \subfigure{
        \includegraphics[width=0.31\textwidth]{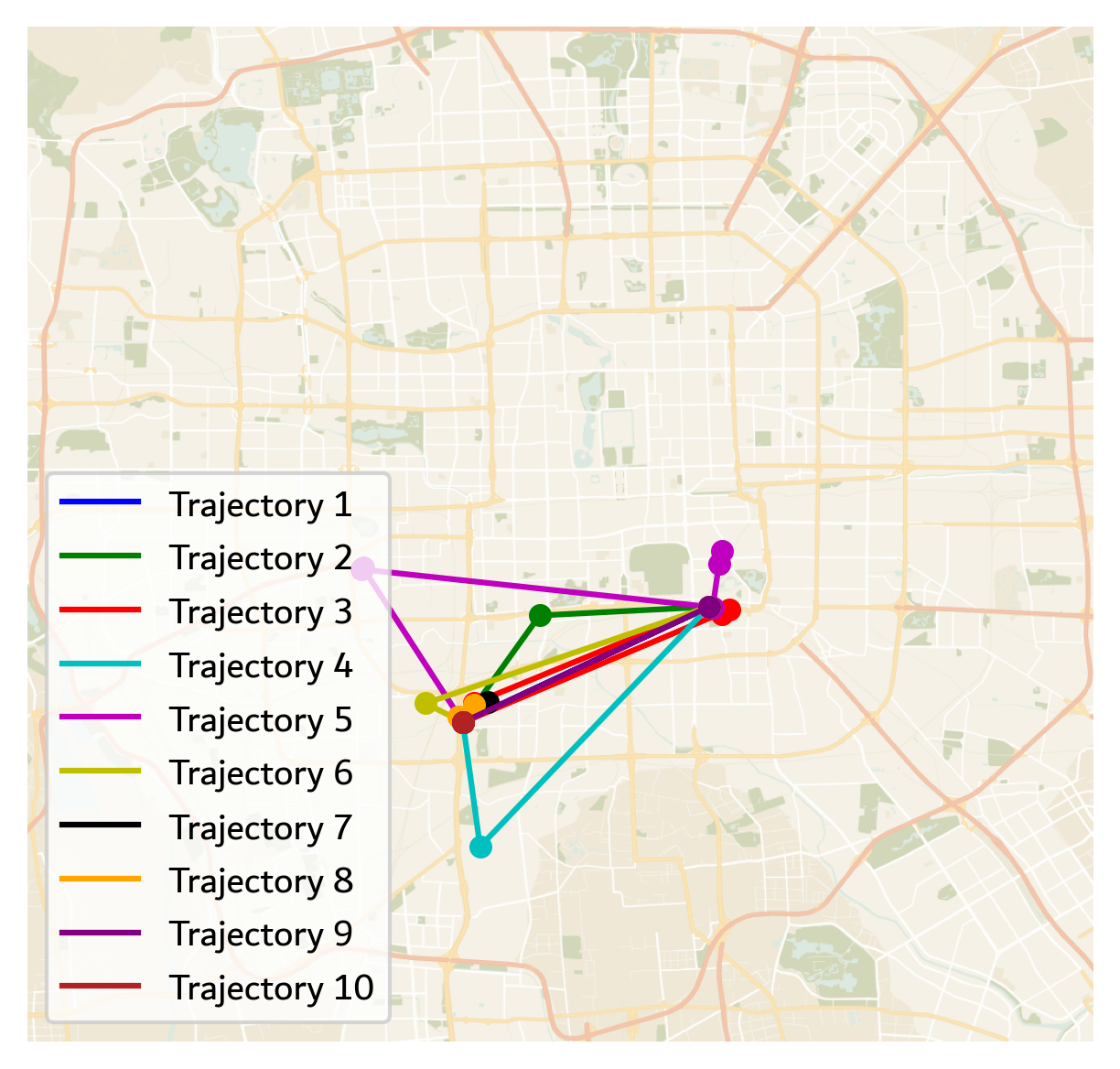}
        \label{fig:sub3}
    }

    \vspace{0.05cm}
    
    \subfigure{
        \includegraphics[width=0.31\textwidth]{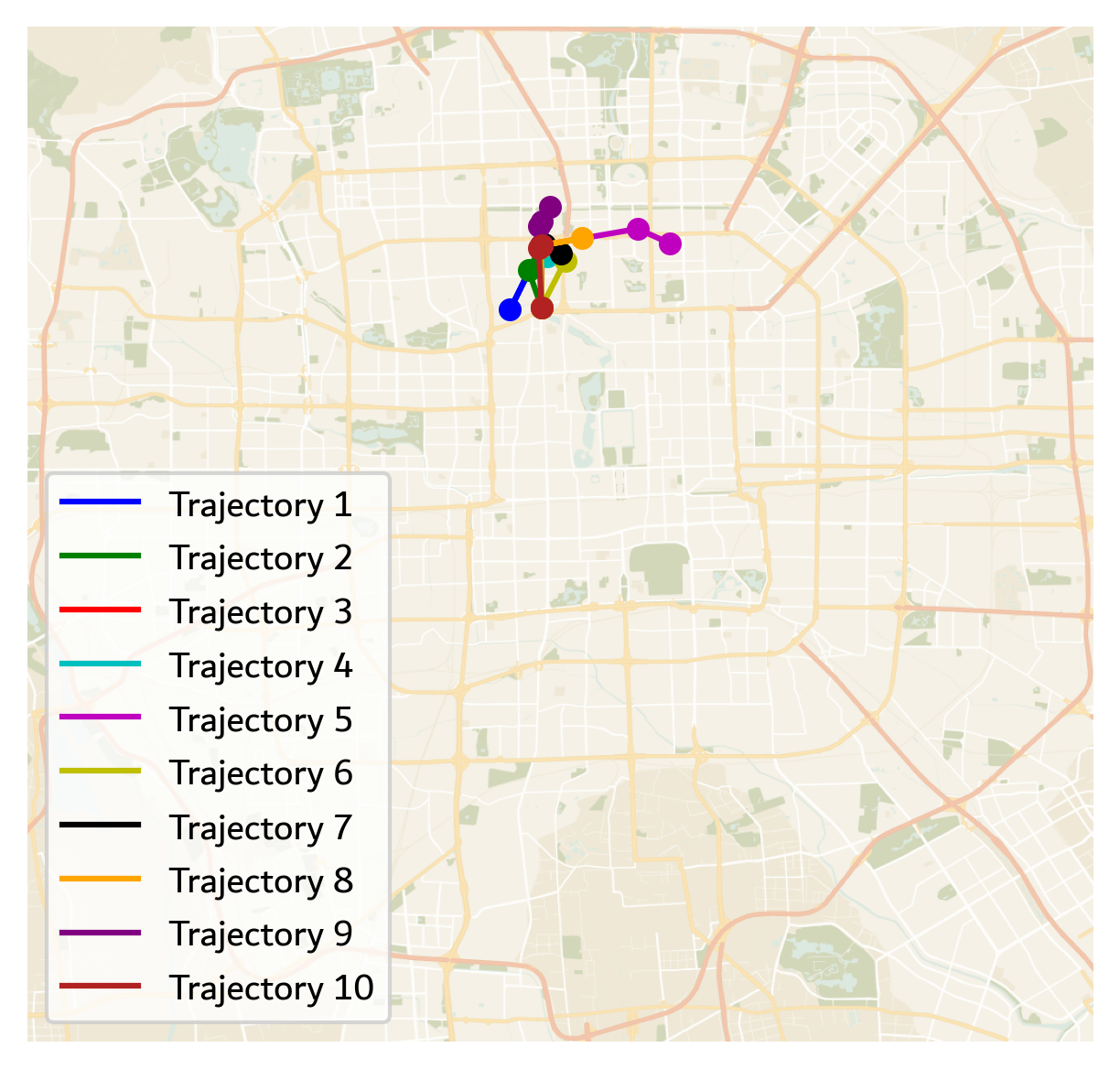}
        \label{fig:sub4}
    }
    \hfill
    \subfigure{
        \includegraphics[width=0.31\textwidth]{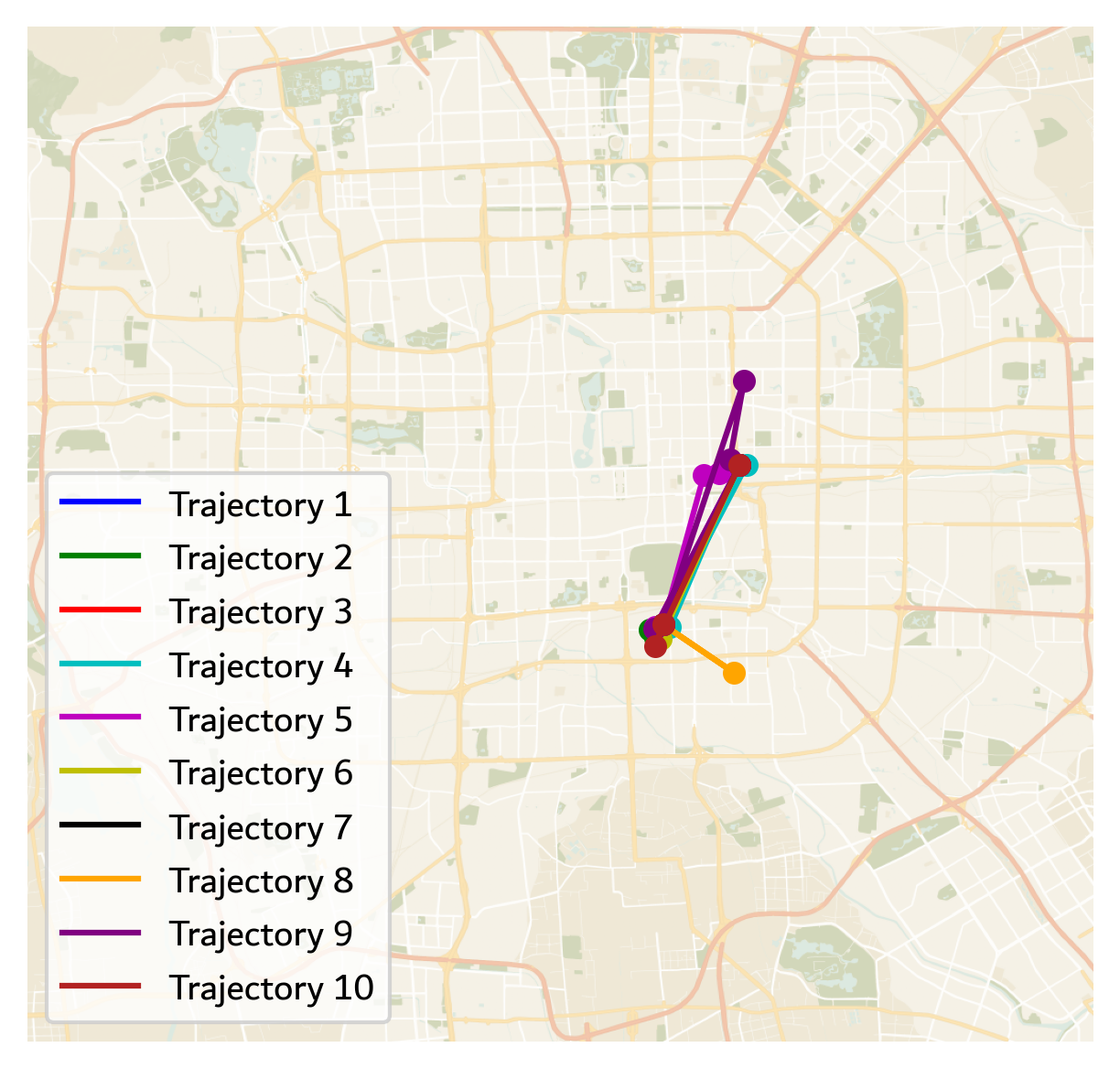}
        \label{fig:sub5}
    }
    \hfill
    \subfigure{
        \includegraphics[width=0.31\textwidth]{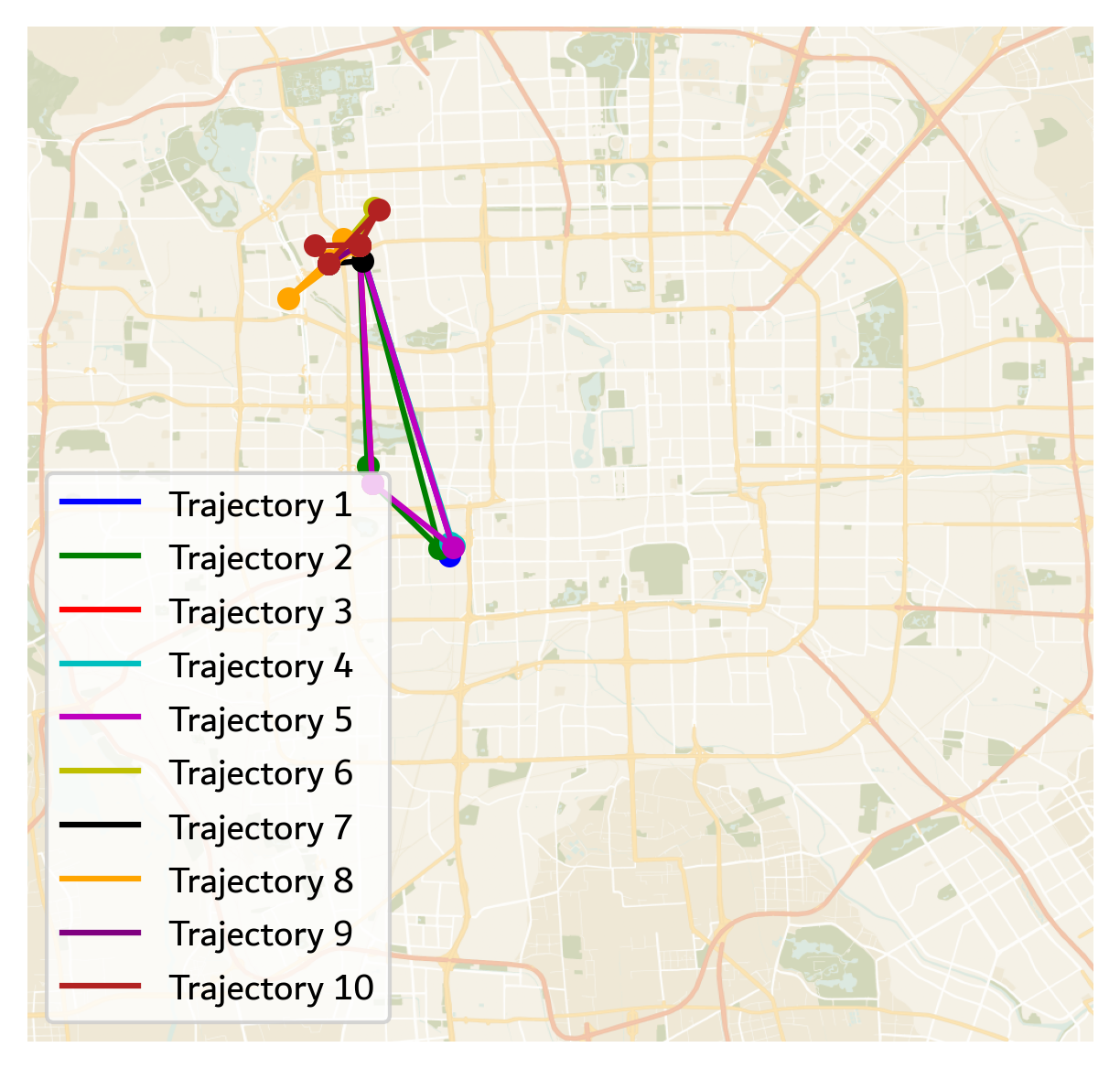}
        \label{fig:sub6}
    }

    \caption{Diversity in geographic space of 6 examples.}
    \label{fig:Diversity-geo}
\end{figure}

The diversity of generated mobile trajectories is a less discussed issue. Previous works have mainly focused on the quality of the data itself, while discussions on diversifying the generated data are lacking. Considering the complexity and diversity of human behaviour, we believe that discussing the diversity of generated trajectories is essential. Moreover, the diversity of trajectories generated by LLM has always been a focal point of concern. Therefore, in this section, we will use additional experiments and visualizations to discuss the diversity issue of the proposed framework. Specifically, we generated 10 1-day results for six people and visualized the trajectories in Figure \ref{fig:Diversity-geo}. Every person's profile, home and work locations are fixed separately.
We use different colors to visualize these 10 different trajectories. As seen in the figure, the locations are mostly distributed around the home and workplace. However, there are also possibilities of visiting more distant POIs.



    


\end{document}